\documentclass[10pt,twocolumn,letterpaper]{article}
\usepackage[accsupp]{axessibility}
\usepackage{amsmath}
\usepackage{algorithm}
\usepackage{algorithmic}
\usepackage{multirow}
\usepackage{verbatim}
\usepackage{gensymb}
\usepackage{subfiles}
\usepackage{amssymb}  

\usepackage[pagenumbers]{cvpr} 

\usepackage[dvipsnames]{xcolor}

\newcommand{\toolname}{READ\xspace}

\definecolor{cvprblue}{rgb}{0.21,0.49,0.74}

\def\paperID{14964} 
\def\confName{CVPR}
\def\confYear{2025}

\usepackage[hidelinks,colorlinks, linkcolor=red, citecolor=black]{hyperref}
\usepackage{graphicx}

\usepackage{booktabs}
\usepackage{tablefootnote}
\usepackage[flushleft]{threeparttable}

\usepackage{lipsum}
\usepackage{bbding}
\usepackage{color}
\usepackage{enumitem}

\usepackage{multirow}
\usepackage{ulem}
\usepackage{tcolorbox}       
\usepackage{amsfonts}       
\usepackage{nicefrac}       
\usepackage{microtype}      
\usepackage{xcolor}
\definecolor{mygreen}{HTML}{3cb44b}

\usepackage[accsupp]{axessibility}

\def\eg{\textit{e.g}\onedot} 
\def\ie{\textit{i.e}\onedot} 
 
\def\etc{\textit{etc}\onedot} 
\def\wrt{w.r.t\onedot} 
\def\aka{a.k.a\onedot}

\title{Reasoning to Attend: Try to Understand How \texttt{<SEG>} Token Works}

\author{Rui Qian$^{1}$
\hspace{0.7cm}Xin Yin$^{3}$
\hspace{0.5cm}Dejing Dou$^{1,2}$\thanks{Corresponding Author 
}
\\
$^{1}$School of Computer Science, Fudan University, $^{2}$BEDI Cloud
~~~\\
$^{3}$The State Key Laboratory of Blockchain and Data Security, Zhejiang University~~~\\
qiianruii@gmail.com, xyin@zju.edu.cn, dejingdou@gmail.com
}

\begin{document}
\maketitle
\begin{figure*}[h]
\vspace{-0.3cm}
\begin{center}
\includegraphics[width=.8\linewidth]{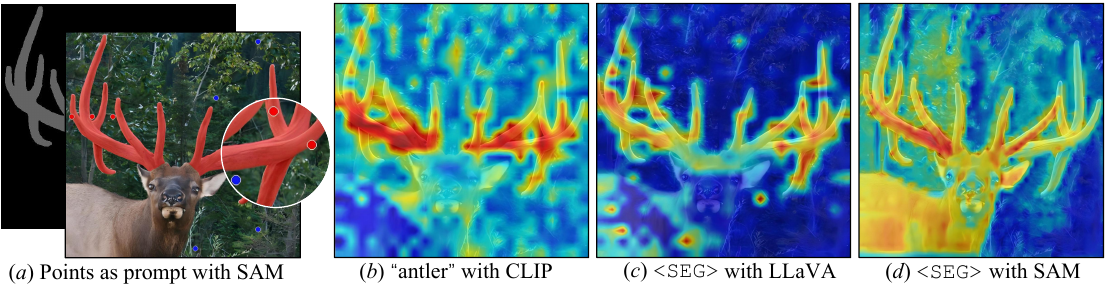}
\end{center}
\vspace{-0.6cm}
\caption{
Qualitative analysis of the 
  \texttt{<SEG>} token on the ReasonSeg \textit{train} set. Points derived from $(c)$ serve as prompts with original SAM in $(a)$. Text ``antler'' with image token from CLIP is in $(b)$.
The similarity between the \texttt{<SEG>} token and image token embeddings stemming from the last hidden layer is obtained by Eq.\eqref{eq:similarity}, \wrt LLaVA encoder in $(c)$ and SAM decoder in $(d)$. The consistency observed in $(b)$, $(c)$, $(d)$ indicates that the \texttt{<SEG>} token in
LMMs learns semantics similar to  direct
mentions in text. Refer to Appendix~\ref{sup:additional_analysis} for more illustrations.}
\label{fig:analysis}
\vspace{-0.55cm}
\end{figure*}

\begin{abstract}
Current Large Multimodal Models (LMMs) empowered visual grounding typically rely on \texttt{<SEG>} tokens as a text prompt to jointly optimize the vision-language model (e.g., LLaVA) and the downstream task-specific model (\eg, SAM). However, we observe that little research has looked into how it works.
In this work, we first visualize the similarity maps, which are obtained by computing the semantic similarity between the \texttt{<SEG>} token and the image token embeddings derived from the last hidden layer in both the LLaVA encoder and SAM decoder. Intriguingly, we have found that a striking consistency holds in terms of activation responses in the similarity map,
which reveals that what the \texttt{<SEG>} token contributes to is semantic 
similarity within image-text pairs. 
Specifically, the \texttt{<SEG>} token, a placeholder expanded in text vocabulary, extensively queries among individual tokenized image patches to match the semantics of an object from text to the paired image, while the Large Language Models (LLMs) are being fine-tuned. Upon the above findings, we present \toolname, which facilitates LMMs' resilient \textbf{REA}soning capability of where to atten\textbf{D} under the guidance of highly activated points borrowed from similarity maps. Remarkably, READ features an intuitive design, Similarity as Points module (SasP), which can be seamlessly applied to \texttt{<SEG>}-like paradigms 
in a plug-and-play fashion.
Also, extensive experiments have been conducted on ReasonSeg and RefCOCO(+/g) datasets. To validate whether READ suffers from 
catastrophic forgetting of previous skills after fine-tuning, we further assess its generation ability on an augmented FP-RefCOCO(+/g) dataset. All codes and models are publicly available at \href{https://github.com/rui-qian/READ}{https://github.com/rui-qian/READ}.
\end{abstract}
    
\vspace{-0.9cm}

\section{Introduction}
\label{sec:intro}

Reasoning segmentation has been newly proposed yet largely unexplored at present \cite{lai2024lisa}. As an extension of classical Referring Expression Segmentation (RES) \cite{hu2016segmentation}, it aims to output nuanced masks for implicitly referred objects given descriptive language expressions. As shown in Fig.~\ref{fig:analysis}, when asked, ``What part of the deer's body in the picture is used for defense and attracting mates?", reasoning segmentation infers ``antler'' without an explicit mention, in contrast to traditional RES, which relies on direct referring. Solving such intricate visual tasks is non-trivial, requiring models to comprehend user intentions based on given queries while also possessing pertinent world knowledge \cite{wu2024see}.

Recent works \cite{lai2024lisa, wu2024see, xia2024gsva, rasheed2024glamm, ren2024pixellm} 
have advanced reason segmentation tasks by 
leveraging \texttt{<SEG>} tokens as a text prompt to seamlessly align LMMs empowered visual encoder (\eg, LLaVA \cite{liu2024visual}) and the downstream task-specific decoder (\eg, SAM \cite{kirillov2023segment}) in vision space for fine-grained output formats, \ie, segmentation masks. Specifically, SESAME \cite{lai2024lisa}
teaches LMMs to respond to false premises
by introducing negative samples into the pipeline. GSVA \cite{xia2024gsva} bridges the gap where the multiple-target and empty-target cases are neglected. GLaMM \cite{rasheed2024glamm} 
and PixelLM \cite{ren2024pixellm} enhances the model both in textual and visual domains, with versatile capability at various levels of granularity. 

However, we observe that little research has looked into how the \texttt{<SEG>} token works when mapping language vocabulary embeddings into corresponding visual space. The \texttt{<SEG>} token, an extended placeholder in the text vocabulary, lacks inherent semantics on its own. Nevertheless, when inserted into conversation templates and jointly trained with LMMs, it becomes capable of grounding objects within an image. Recent works \cite{lai2024lisa,wu2024see, xia2024gsva} all employ the SAM \cite{kirillov2023segment} model as a mask decoder. Initially, segmenting the red region in Fig.~\ref{fig:analysis} could be achieved by prompting SAM with the text ``antler", but now the \texttt{<SEG>} token embedding fulfills the same purpose in place. This leads us to ask: \textit{What is the connection of embeddings between the \texttt{<SEG>} token and the text prompt ``antler'' in terms of semantics?}

Bearing this in mind, we begin by visualizing similarity maps, which are generated by computing dot product similarity between the \texttt{<SEG>} token and the image token embeddings extracted from the last hidden layer of both LLaVA \cite{liu2024visual} and SAM \cite{kirillov2023segment} models. Notably, 
we observe a striking consistency in activation responses across  similarity
maps, suggesting that the \texttt{<SEG>} token is pivotal in bridging semantic connections between textual prompts and their visual correspondences. Specifically, the \texttt{<SEG>} token, an expansion of text vocabulary, initiates a thorough query across each tokenized image patch, aligning the textual semantics of an object with its visual counterparts in the paired image. Inspired by the above findings, an intuitive idea is to see if we can imply to the model where to ``attend" by leveraging similarity maps.

To this end, we present READ, which 
facilitates LMMs' resilient \textbf{REA}soning capabilities of where to atten\textbf{D}, 
informed by highly activated points stemming from similarity
maps. In particular, our READ consists of three core modules: (1) a LLaVA encoder, (2) a Similarity as Points module (SasP), and (3) a SAM decoder. Specifically, our LLaVA \cite{liu2024visual} enhanced encoder consumes image-text pairs as input to generate text output, from which 
the last hidden layer embedding for the  \texttt{<SEG>} token is then gathered. To guide the model where to ``attend'', our SasP module computes similarity maps by performing a dot product between the \texttt{<SEG>} token embedding and the associated image patches, whereupon regions with high similarity scores are then converted into point coordinates for fine-grained mask predictions through the SAM decoder, along with textual prompts, \ie the \texttt{<SEG>} token embedding. To address the challenge posed by discrete, non-differentiable points during back-propagation, we apply a Gaussian-like weighted average interpolation to render them continuously differentiable. This modification facilitates gradients across similarity maps back to the LMMs, empowering the model to ``reason to attend'' in the forward, and ``attend to reason'' in the backward and vice versa. Particularly, READ’s intuitive design, SasP, can be effortlessly integrated into off-the-shelf \texttt{<SEG>}-like paradigms with minimal overheads in a plug-and-play manner. In summary, our contributions are threefold: 
\vspace{-0.1cm}
\begin{itemize}[leftmargin=0.3cm]
    \item We have looked into how the \texttt{<SEG>} token works when mapping
language vocabulary embedding into corresponding visual space. Such investigation reveals that what the  \texttt{<SEG>} token mainly contributes to is the semantic correspondences from image-text pairs based on our findings.
    \item We present our model --- READ, 
    which empowers LMMs to “reason to attend” and “attend to reason”, vice versa. Importantly, our intuitive and simple design, SasP, can be effortlessly integrated into off-the-shelf \texttt{<SEG>}-like pipelines with marginal overheads.
    
    \item We conduct extensive experiments on both the challenging reasoning segmentation dataset and the well-established RefCOCO(+/g) referring segmentation dataset. To see if READ struggles with catastrophic forgetting of previous capabilities, we also assess its generative performance on the FP-RefCOCO(+/g) dataset. Our READ surpasses the existing state-of-the-art by a 
    remarkable margin, resulting in cIoU improvements over baselines of up to $4.7\%$ on ReasonSeg, $3.73\%$ on FP-RefCOCO(+/g).
\end{itemize}

\vspace{-0.19cm}
\section{Related Work}
\vspace{-0.2cm}
\subsection{Large Multimodal Models}
\vspace{-0.2cm}
Depending on the level of capabilities 
LMMs possess, we categorize them into three  groups: (1) \textit{Multimodal feature alignment},  which aligns visual and textual features toward a comprehensive multimodal understanding~\cite{alayrac2022flamingo, li2023blip, ye2024mplug}. Flamingo~\cite{alayrac2022flamingo} enables visual in-context reasoning through cross-attention. BLIP-2~\cite{li2023blip} utilizes a lightweight query-based visual encoder to align image features with a frozen language model. mPLUG-OWL~\cite{ye2024mplug} connects image encoders to language models via efficient prompts. (2) \textit{Instruction tuning for few-shot learning}, which leverages instruction tuning to enable few-shot learning capabilities, allowing models to acquire new skills even with limited samples \cite{li2023otter, liu2024visual, zhu2023minigpt}. Otter \cite{li2023otter} enhances LMMs with instruction-tuning on the proposed MIMIC-IT dataset. LLaVA~\citep{liu2024visual} and MiniGPT-4~\citep{zhu2023minigpt} integrate a visual encoder for feature extraction and align image representations with text embeddings, effectively enhancing their capabilities in a variety of vision-language tasks. (3) \textit{Fused task and enhanced reasoning}, which advances LMMs' reasoning capabilities with a unified interface for versatile vision-centric tasks \cite{wang2023visionllm, detgpt, peng2023kosmos, zhang2023gpt4roi}. VisionLLM \citep{wang2023visionllm} unifies diverse visual tasks within a single language model interface. Kosmos-2~\citep{peng2023kosmos} and DetGPT~\citep{detgpt} inject grounding capabilities into LMMs. GPT4RoI~\cite{zhang2023gpt4roi} enables region-level understanding by constructing region-text pairs. In contrast, our READ builds upon LLaVA~\cite{liu2024visual} for 
world knowledge and complex reasoning.

\vspace{-0.1cm}
\subsection{Interactive Segmentation Models}
\vspace{-0.1cm}
Depending on whether the interactive capabilities are present in the models, 
we classify literature into two groups: 
(1) \textit{ Non-interactive segmentation}, which 
assigns class labels to the pixel-level in a scene (\ie, semantic segmentation \cite{long2015fcn, ronneberger2015unet, badrinarayanan2017segnet}), or 
object-level of a scene (\ie, instance segmentation \cite{he2017mask, liu2018path, carion2020DETR}), or both of which simultaneously (\ie, panoptic segmentation \cite{kirillov2019panopticps,xiong2019upsnet,kirillov2019panopticfpn}). 
U-Net \cite{ronneberger2015unet} features an encoder-decoder architecture that incorporates skip connections to preserve feature information. Mask R-CNN \cite{he2017mask} introduces an additional segmentation branch to Faster R-CNN \cite{ren2016faster}, allowing for object detection and instance segmentation in parallel. PS \cite{kirillov2019panopticps} initially introduces the concept of panoptic segmentation, defining it as the task of labeling every pixel in an image, including both segmentable objects (like people and vehicles, \etc) and unsegmentable ``stuff'' classes (such as sky and road, \etc). (2) \textit{Interactive segmentation}, which aims to interact with human language, enabling models to segment target objects based on descriptive texts, typically, Referring Expression Segmentation (RES) \cite{xia2024gsva, rasheed2024glamm, ren2024pixellm, lai2024lisa, wu2024see}. GSVA \cite{xia2024gsva} boosts RES by enabling the identification of multiple objects from a single description and recognizing absent targets. GLaMM \cite{rasheed2024glamm} and PixelLM \cite{ren2024pixellm} augment models' capabilities in both textual and visual domains, showcasing versatility across multiple levels of granularity. Closest to our work, LISA \cite{lai2024lisa} and 
SESAME \cite{wu2024see} inject self-reasoning capabilities into segmentation tasks, elevating RES into even more advanced interactions, \ie, reasoning segmentation. 

Note that aforementioned literature, \wrt RES leans on the \texttt{<SEG>} token as the intermediate connector to link the downstream mask decoder. Whereas, we observe that few investigations have looked into how it works so far. This work aims to unveil how the \texttt{<SEG>} token contributes to, whereupon we present the proposed method, READ.

\section{Reflection on Reasoning Segmentation}
In this section, we first revisit the reasoning segmentation task and then analyze the underlying mechanisms of how the \texttt{<SEG>} token works upon the prior state-of-the-art \cite{lai2024lisa, wu2024see}.
\vspace{-0.2cm}
\subsection{Revisiting}
\begin{figure*}[t]
\begin{center}
\includegraphics[width=1\linewidth]{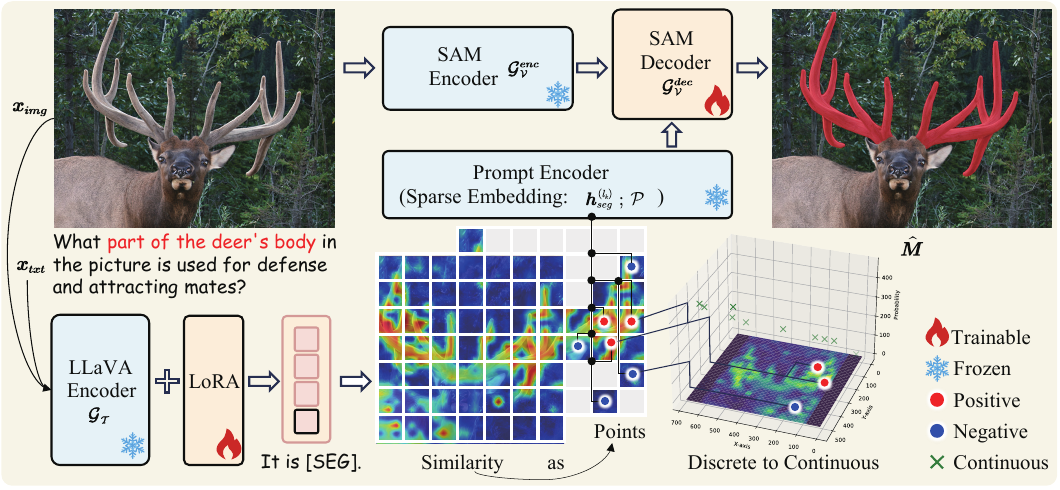}
\end{center}
\vspace{-0.5cm}
\caption{
Overview of our proposed READ. The hidden state outputs with respect to the \texttt{<SEG>} token and image tokens are derived from the LLaVA encoder for similarity as points, before being fed into the prompt encoder for sparse embedding. 
To inform the model where to ``attend'' when reasoning, we apply
a Gaussian-like weighted average interpolation to transform discrete points into continuous ones.}
\label{fig:overview}
\vspace{-0.4cm}
\end{figure*}

\vspace{-0.1cm}
\textbf{Problem Definition:}
Let $\mathbf{x}_{img} \in \mathbb{R}^{h \times w \times c}$ denote the input image, where $h$, $w$ and $c$ are the height, width and channels of the image, respectively. 
Consider the paired textual input, denoted by $\mathbf{x}_{txt}$, which can range from an explicit mention, such as ``antler" to an implicit expression like ``part of the deer’s body". Reasoning segmentation primarily involves the task of generating a segmentation mask $\hat{\mathbf{M}}$ such that it aligns with the part of the image that corresponds to the referenced query as 
\begin{align}
\vspace{-0.2cm}
\begin{aligned}
\Theta _{MLE}=\underset{\Theta}{\mathrm{arg}\max}\mathcal{G} _{\theta}\left( \hat{\mathbf{M}}|\mathbf{x}_{img},\mathbf{x}_{txt};\Theta \right),
\end{aligned}
\vspace{-0.2cm}
\end{align}
where $\mathcal{G}_{\theta}$ indicates segmentation capabilities infused into LMMs parameterized by $\Theta$. 
In this paper, it includes a multi-modal 
LLM $\mathcal{G} _{\mathcal{T}}$ and a visual backbone model $\mathcal{G} _{\mathcal{V}}$. Concisely, $\mathcal{G} _{\theta}=\mathcal{G} _{\mathcal{T}}\oplus \mathcal{G} _{\mathcal{V}}$, $\oplus$ denotes the cascading operation. As illustrated in Fig.~\ref{fig:overview}, $\mathcal{G} _{\mathcal{T}}$ and $\mathcal{G} _{\mathcal{V}}$ are instantiated by 
LLaVA \cite{liu2024visual} and SAM \cite{kirillov2023segment} accordingly.
$\hat{\mathbf{M}} \in \{0, 1\}^{h \times w}$, $1$ indicates the presence of an object and $0$ otherwise. 

To facilitate $\mathcal{G}_{\mathcal{\theta}}$ in an embedding as mask fashion, LISA \cite{lai2024lisa} 
extends the text vocabulary of $\mathcal{G}_{\mathcal{T}}$ with a placeholder, \ie, the \texttt{<SEG>} token. To enable LLMs to tackle image features in the same way as text sequences, $\mathbf{x}_{img}$ is segmented into patches of equal size and transformed by the CLIP \cite{radford2021learningclip} model. During training, the \texttt{<SEG>} token is embedded within $\mathbf{x}_{txt}$, and then both $\mathbf{x}_{txt}$ and $\mathbf{x}_{img}$ are fed into LMMs $\mathcal{G}_{\mathcal{T}}$, which in turn generates a text response $\hat{\mathbf{y}}_{txt}$ as
\vspace{-0.4cm}
\begin{align}
\begin{aligned}
\hat{\mathbf{y}}_{txt}=\;\mathcal{G}_{\mathcal{T}} (\mathbf{x}_{img},\mathbf{x}_{txt}).
\end{aligned}
\end{align}
During inference, when $\mathcal{G}_{\theta}$ is prompted interactively to output a binary segmentation mask, the response $\hat{\mathbf{y}}_{txt}$ would include a \texttt{<SEG>} token if the object exists. LISA \cite{lai2024lisa} then retrieves the last hidden layer embedding $\tilde{\boldsymbol{h}}_{seg}$
from $\mathcal{G}_{\mathcal{T}}$  associated with the predicted \texttt{<SEG>} token, which is then  passed through a multilayer perceptron (MLP) projection layer, denoted by $\mathrm{\varphi}$ to obtain the refined feature $\boldsymbol{h}_{seg}$. Concurrently, the vision backbone  $\mathcal{G} _{\mathcal{V}}^{enc}$ borrowed from SAM \cite{kirillov2023segment} is employed to extract a rich set of visual features $\mathbf{f}$ from the visual input $\mathbf{x}_{img}$, represented by 
\begin{align}
\begin{aligned}
\boldsymbol{h}_{seg} = \varphi(\tilde{\boldsymbol{h}}_{seg}), \quad \mathbf{f} = \mathcal{G} _{\mathcal{V}}^{enc}(\mathbf{x}_{img}). \label{eq:h_seg}
\end{aligned}
\end{align}
At last, visual features $\mathbf{f}$ are fed into mask decoder $\mathcal{G} _{\mathcal{V}}^{dec}$ to produce the final segmentation mask $\hat{\mathbf{M}}$ conditioned by embedding $\boldsymbol{h}_{seg}$ as 
\begin{align}
\begin{aligned}
    \hat{\mathbf{M}} = & \; \mathcal{G} _{\mathcal{V}}^{dec}(\mathbf{f}, \boldsymbol{h}_{seg}).
    \label{eq:hatmask}
\end{aligned}
\end{align}
$\boldsymbol{h}_{seg}$ importantly bridges the intermediate layers, informing the mask decoder to seamlessly decode the segmentation mask in a way of end-to-end training. This paradigm has been widely adopted by its successors \cite{wu2024see,rasheed2024glamm,ren2024pixellm,xia2024gsva}. However, we observe that few investigations have looked into how the embedding $\boldsymbol{h}_{seg}$ works so far, which inspires us to explore its underlying mechanism.

\subsection{Analysis}
\label{sec:Analysis}
To qualitatively analyze the \texttt{<SEG>} token, we visualize the similarity maps at different stages of the forward pass through $\mathcal{G}_{\theta}$ in SESAME \cite{wu2024see}. As illustrated in Fig. \ref{fig:analysis}, ``\texttt{<SEG>} with LLaVA'' denotes the similarity map between the embedding of the \texttt{<SEG>} token, \ie, $\boldsymbol{h}_{seg}$ and the image tokens from the last hidden layer outputs. Activation responses in $(c)$ and $(d)$ indicate that $\boldsymbol{h}_{seg}$ 
signifies the model where to ``attend'' somehow. Further, the consistency between the striking cues in $(b)$ and $(c)$ reveals that $\boldsymbol{h}_{seg}$ plays a key role in aligning with the semantics when bridging the intermediate layers. $(b)$ is obtained by using the embedding of text ``antler'' and visual features from CLIP \cite{radford2021learningclip}, which indirectly suggests that the \texttt{<SEG>} token has acquired semantics akin to the direct mention of ``antler'', when prompted by the implicit expression “part of the deer's body”.

To quantitatively assess the effects of the \texttt{<SEG>} token, we conduct experiments focusing on similarity maps. We first select several points with the highest similarity scores as positives and an equal number of points with the lowest similarity scores as negatives. These points are then directly used as prompts instead of the \texttt{<SEG>} token and are input into the original SAM model to generate the segmentation mask (see Appendix~\ref{fig:Appendix02}).
%
In Table~\ref{tab:Analysis}, $\mathcal{P}$\textsubscript{prompt} reveals that relying solely on the selected similarity points can still potentially generate a segmentation mask ($27.0\%$ vs. $30.4\%$). 
To further quantify the extent to which the activated points within the similarity map in $(c)$ correspond to the target object in $(a)$, we adopt the grid search-based Intersection over Union (see Appendix~\ref{sup:GridSearchforOptimal Threshold}) to validate the consistency between the similarity map and the ground-truth mask. In Table~\ref{tab:Analysis}, 
$\mathcal{S}$\textsubscript{IoU} suggests that responses in  similarity maps are strikingly consistent with the ground-truth mask, with a 6$\%$ higher cIoU on the ReasonSeg \textit{test} set (36.4$\%$vs.30.4$\%$) in the $2^{nd}$ row.
\vspace{-0.1cm}
\begin{table}[htbp]
  \centering
  \caption{
  Quantitative analysis of the 
  \texttt{<SEG>} token on the ReasonSeg \textit{test} set. $\mathcal{P}$\textsubscript{prompt} denotes points as prompt for original  SAM~\cite{kirillov2023segment}. $\mathcal{S}$\textsubscript{IoU} measures the overlap (IoU) between the similarity map and the ground-truth mask. \texttt{<SEG>}\textsubscript{prompt} denotes the  \texttt{<SEG>} token as prompt for the adapted SAM, \aka  LISA~\cite{lai2024lisa}.}
    \vspace{-0.2cm}
\resizebox{.9\linewidth}{!}
{
    \begin{tabular}{l|cc|cc|cc}
    \toprule
    \multirow{2}[4]{*}{Method} & \multicolumn{2}{c|}{$\mathcal{P}$\textsubscript{prompt}} & \multicolumn{2}{c|}{$\mathcal{S}$\textsubscript{IoU}}& \multicolumn{2}{c}{\texttt{<SEG>}\textsubscript{prompt}} \\
\cmidrule{2-7}          & gIoU  & cIoU  & gIoU  & cIoU & gIoU  & cIoU \\
    \midrule
    LISA-7B \cite{lai2024lisa}  & \textbf{38.2}  & \textbf{30.1}  & 32.1  & 34.1 & \textbf{47.3} & \textbf{48.4} \\
    SESAME \cite{wu2024see} & 35.9  & 27.0    & \textbf{32.5}  & \textbf{36.4} & 30.5 & 30.4\\
    \bottomrule
    \end{tabular}%
}
 \label{tab:Analysis}%
 \vspace{-0.1cm}
\end{table}%

\noindent \textbf{Summary}. By analyzing the effects of the \texttt{<SEG>} token, we observe the following: $(a)$ The \texttt{<SEG>} token in LMMs learns semantic features similar to those of direct mentions in text  and aligns these textual semantics with its visual space to guide the generation of the segmentation mask. $(b)$ The activated points within similarity maps imply the locations of the target object, which to some extent provides feedback on where the model is focusing. Understanding how the \texttt{<SEG>} token works is crucial as it is closely tied to the issue of semantic alignment within LLMs. The \texttt{<SEG>} token offers insights into the experimental observation that, when prompting LISA~\cite{lai2024lisa} for reasoning explanations, the textual outputs from the LLaVA encoder remain accurate, even for cases where the SAM decoder fails in segmentation. These reflections motivate us to directly leverage similarity points 
to guide the model where to ``attend'' when reasoning. For further analysis, please see the Appendix~\ref{sup:additional_analysis}.

\section{Proposed READ}

In this section, we present READ, which unlocks LMMs' resilient reasoning capability of where to
``attend'' under the guidance of highly activated points derived from similarity maps. As shown in Fig. \ref{fig:overview}, the proposed READ includes: $(a)$ an LLaVA encoder, $(b)$ a Similarity as Points Prompter (SasP), and $(c)$ a SAM  decoder. In READ,  we first use the LLaVA encoder to take image-text pairs as input, which in turn responds to text as outputs. We then extract the embedding of  the \texttt{<SEG>} token and image tokens from the last hidden layer of the LLaVA encoder to compute the similarity map, upon which we employ our Discrete to Continuous sampling (DtoC) to convert highly activated foreground points into continuous ones. Finally, these continuous points along with the \texttt{<SEG>} token embedding are fed into the SAM \cite{kirillov2023segment}  decoder for mask generation. 
As these points are differentiable, the loss will be backpropagated to LMMs to signify $\mathcal{G}_{\mathcal{T}}$ ``where to attend'' when reasoning.
Given that our innovation mainly lies in SasP, we discuss it first in Sec.~\ref{method:arch}.

In what follows, we present the LLaVA encoder in Sec.~\ref{LLaVAEncoder}. 
SAM~\cite{kirillov2023segment} decoder in Sec.~\ref{SAMMaskDecoder}, and the Training objectives of READ are detailed in Sec.~\ref{TrainingObjectives}.

\subsection{Similarity as Points}
\label{method:arch}
\paragraph{Points as Prompt.} 
Inspired by various types of prompts supported by SAM \cite{kirillov2023segment} mask decoder, such as bounding boxes, points, and dense mask, \etc, we explore how to derive the points of interests as prompts in the input image $\mathbf{x}_{img}$ via the similarity score. Specifically, 
denote $\boldsymbol{h}^{\left( l_k \right)}=\left\{ \boldsymbol{h}_{1}^{\left( l_k \right)},...,\boldsymbol{h}_{N_k}^{\left( l_k \right)}|\boldsymbol{h}_{i}^{\left( l_k \right)}\in \mathbb{R} ^d \right\} 
$ as the hidden state output at the $k$-th layer of $\mathcal{G}_{\mathcal{T}}$, where $N_k$ denotes the number of hidden state tokens, and $d$ is the embedding dimension. $\boldsymbol{h}_{seg}$ in Eq.~\eqref{eq:h_seg} can be represented as $\boldsymbol{h}_{seg}^{\left( l_k \right)}\in \boldsymbol{h}^{\left( l_k \right)}$. In the course of training, the image features are incorporated into the text instruction 
$\mathbf{x}_{txt}$ and fed as input to the LLM. Hence, the hidden state output at the $k$-th layer, $\boldsymbol{h}^{\left( l_k \right)}$, includes $N_t$ image tokens, denoted as
$\boldsymbol{h}_{img}^{\left( l_k \right)}\subseteq \boldsymbol{h}^{\left( l_k \right)}$. We formulate the similarity score between \texttt{<SEG>} token and image tokens as
\vspace{-0.3cm}
\begin{align}
\begin{aligned}
\mathcal{S} =\boldsymbol{h}_{img}^{\left( l_k \right)}\cdot \left( \boldsymbol{h}_{seg}^{\left( l_k \right)} \right) ^{\mathrm{T}},
\label{eq:similarity}
\end{aligned}
\end{align} where $\mathcal{S}$ denotes similarity score between each image token and the \texttt{<SEG>} token,
$\mathcal{S} \in \mathbb{R} ^{N_t}$. Note that our vanilla similarity score is parameter-free, which could be easily amenable to
\texttt{<SEG>}-like paradigms with negligible effort.
Since this paper primarily aims to explore how the \texttt{<SEG>} token works, 
we leave designing possibly more effective similarity computation strategies for  READ as future work, such as using cross attention \cite{vaswani2017crossattention} to acquire learnable fine-grained patterns.
Nevertheless, Eq.~\eqref{eq:similarity} is already sufficient to generate the necessary cues according to $(c)$ in Fig.~\ref{fig:analysis}. 

To facilitate similarity score as point prompt, we restore the coordinates over three types of points in $\mathbf{x}_{img}$, \ie, positive, negative, and neutral points. Positive points denote those that are confidently associated with an object or the foreground. Negative points denote those that are confidently associated with the background. Neutral points denote those that are not clearly identifiable as belonging to either the object or the background, which could be near object boundaries, ambiguous regions, or areas.
To this end, Algorithm~\eqref{alg:sasp} outlines the procedure:
$\left( \mathrm{i} \right)$ In Steps 1-3, we normalize the similarity score $\mathcal{S}$ and set the thresholds for positive and negative points based on the mean $\mu$ and variance $\sigma$ of $\mathcal{S}$. $\left( \mathrm{ii} \right)$ In Steps 6-7, we recover each 
selected point $j$ with the absolute coordinates. $\left( \mathrm{iii} \right)$ In Steps 8-10, since the point selection process involves operations such as sorting, $j$ becomes a discrete, non-differentiable value. To allow for gradient backpropagation, we apply a Gaussian-like weighted average interpolation to obtain continuous, differentiable coordinates. The weights are computed based on the distance to each grid point. After that, point set $\mathcal{P}$ along with the \texttt{<SEG>} token embedding $\boldsymbol{h}_{seg}^{\left( l_k \right)}$ is fed into $\mathcal{G} _{\mathcal{V}}^{dec}$ as input, we reformulate Eq.~\eqref{eq:hatmask} as
\begin{align}
\begin{aligned}
\hat{\mathbf{M}}=\;\mathcal{G} _{\mathcal{V}}^{dec}(\mathbf{f},  \boldsymbol{h}_{seg}^{\left( l_k \right)}, \mathcal{P} ).
\end{aligned}
\end{align}
The shift from discrete points to continuous, differentiable ones is crucial, as 
the gradients propagated backward during training are expected to guide the model in refining its focus. If the similarity score of background points is higher, causing positive points to fall within the background, it will degrade the mask result and increase the corresponding loss, which in turn, penalizes the model and encourages it to learn where to ``attend''. Considering that Steps 8-10 form the core of SasP, we go through them in detail.
\vspace{-0.3cm}
\paragraph{Discrete to Continuous.}
Let $\mathbf{g}_{\boldsymbol{x}}$ and $\mathbf{g}_{\boldsymbol{y}}$ represent the coordinates of grid points, and $\left( x_j, y_j \right)$ denote the coordinates of the selected point $j$. 
By using distance-based weights and normalized softmax probabilities, the sampling process will be continuously differentiable with respect to selected point $\left( x_j, y_j \right)$ in the limit as the grid resolution increases.
Specifically, the weight for each grid point is computed based on the distance to the selected point, using an exponential decay function. This ensures that closer points have higher weights, and farther points have lower weights as
\begin{align}
\vspace{-0.2cm}
\begin{aligned}
w_{i}^{j}=\exp \left( -d_{i}^{j} \right),
\end{aligned}
\vspace{-0.2cm}
\end{align}
where $d_{i}^{j}$ denotes the distance between grid point $i$ and the selected point $j$.
To incorporate both the distance weight and the softmax probability $p_i$ (from the similarity map $\mathcal{S}$), the final weight for each grid point is computed as
\begin{align}
\vspace{-0.3cm}
\begin{aligned}
\tilde{w}_{i}^{j}=w_{i}^{j}\cdot p_i,    p_i=\small{\frac{\exp \left( \mathcal{S} _i \right)}{\Sigma _{i^{\prime}}\exp \left( \mathcal{S} _{i^{\prime}} \right)}}.
\end{aligned}
\end{align}
These weights are normalized so that their sum equals 1, ensuring that they form a valid probability distribution as
\vspace{-0.4cm}
\begin{align}
\begin{aligned}
\hat{w}_{i}^{j}=\frac{\tilde{w}_{i}^{j}}{\Sigma _{i^\prime}\tilde{w}_{i^\prime}^{j}},
\end{aligned}
\end{align}
where the sum is taken over all grid points $i^\prime$. Finally, the continuous coordinates $\left( \hat{x}_i,\hat{y}_j \right) $ for each selected point $\left( x_j,y_j \right)$ 
are computed by taking a weighted average of the grid coordinates, using the normalized final weights $\hat{w}_{i}^{j}$ as
\begin{align}
\begin{aligned}
\hat{x}_j=\underset{i=1}{\overset{h\times w}{\Sigma}}\mathbf{g}_{x,i}\cdot \hat{w}_{i}^{j},  \hat{y}_j=\underset{i=1}{\overset{h\times w}{\Sigma}}\mathbf{g}_{y,i}\cdot \hat{w}_{i}^{j},
\end{aligned}
\end{align}
where $\mathrm{g}_{x,i}$ and $\mathrm{g}_{y,i}$ are the $x$-th and $y$-th coordinates of grid point $i$, respectively. Given that the weight function $\exp \left( -d_{i}^{j} \right) $ decays smoothly with the distance between $\mathrm{g}_i$ and $\left( x_j, y_j \right)$, $\hat{w}_{i}^{j}$ forms a smooth probability distribution over the grid points. Thereby, the weighted sum of grid points converges to continuous interpolation as the grid is refined, \ie, in the limit as the grid resolution
$\varDelta x\rightarrow 0$ (the distance between grid points approaches zero), the discrete weighted sum can be approximated by an integral over a continuous domain. The weighted average becomes
\begin{align}
\begin{aligned}
\hat{x}_j=\int{\mathbf{g}_x\left( \mathbf{g} \right) \cdot w\left( \mathbf{g},x_j,y_j \right) \,\,d\mathbf{g},}
\end{aligned}
\end{align}
where $\mathbf{g}$  is a continuous variable representing the position on the grid, and $w\left( \mathbf{g},x_j,y_j \ \right)$ is a smooth weight function based on the distance to the selected point $\left( x_j, y_j \right)$.

\begin{algorithm}[!t]
    \caption{Similarity as Points Algorithm}
    \label{alg:sasp}
    \begin{algorithmic}[1]
        \REQUIRE ~~\\
        $\mathcal{S}$ is the similarity score between each image token and the \texttt{<SEG>} token obtained in Eq.~\eqref{eq:similarity}, where 
        $\mathcal{S} \in \mathbb{R}^{N_t}$, $N_t$ denotes the number of image tokens;
        \\
        $\mathcal{I} _+ $ is the indices of positive points determined by a threshold $t_{pos}$,$\mathcal{I} _- $ is the indices of negative points with a threshold of $t_{neg}$, $\mathcal{I} _0 $ stands for the indices of neutral points.
        $\mathcal{I} =\mathcal{I} _+\cup \mathcal{I} _-\cup \mathcal{I} _0$;
        \\
        $\mathbf{g}$ is the set of grid points, $\mathbf{g}=\left[ h \right] \times \left[ w \right] $, where $\left[ h \right] =\left\{ 1,2,...,h \right\}$ and 
        $\left[ w \right] =\left\{ 1,2,...,w \right\}$. $hw$ denote the raw image $\mathbf{x}_{img}$ dimension. 
        $\mathbf{g}_{\boldsymbol{x}}$ and $\mathbf{g}_{\boldsymbol{y}}$ are the coordinates of grid points along $x$-axis, $y$-axis.
        \ENSURE ~~\\ 
        Selected points and labels:
        $\mathcal{P} =\emptyset$,
        $\mathrm{labels}=\left[ -1 \right] ^{\left| \mathcal{I} \right|}$
                
        \STATE $\mathcal{S} _{\left[ 0,1 \right]}\gets \mathcal{S} , \mathrm{\mu}=\mathbb{E} \left[ \mathcal{S} _{\left[ 0,1 \right]} \right] , \mathrm{\sigma}^2=\mathrm{Var}\left[ \mathcal{S} _{\left[ 0,1 \right]} \right] $

        \STATE $t_{pos} = \mu + \sigma \cdot \mathrm{\varepsilon}$, $t_{neg} = \mu - \sigma \cdot \mathrm{\varepsilon}, \mathrm{\varepsilon}=0.5$
        
        \STATE
        $\mathcal{I} _+ = \left\{ j\ |\ \mathcal{S} _j\ge t_{pos} \right\}$, $\mathcal{I} _-=\left\{ j\ |\mathcal{S} _j \le \ t_{neg}  \right\}$

        \STATE $\alpha =\small{\frac{w}{\lfloor \sqrt{N_t} \rfloor}}, \beta =\small{\frac{h}{\lfloor \sqrt{N_t} \rfloor}}$
        \FOR{each index $j$ in $\mathcal{I}$}
            \STATE $x_j=\min\mathrm{((}j\,\,\mathrm{mod} \ w+0.5)\cdot \alpha ,w-1)$
            \STATE $y_j=\min\mathrm{((}j\div w+0.5)\cdot \beta ,h-1)$
            \STATE $d_{i}^{j}\!=\!\left\| \left( \mathbf{g}_{x,i},\mathbf{g}_{y,i} \right) -\left( x_j,y_j \right) \right\| _2,\forall i\in \mathbf{g}$
            \STATE $\hat{w}_{i}^{j}=\small{\frac{\exp \left( -d_{i}^{j} \right) \cdot p_i}{\Sigma _{i^{\prime}}\exp \left( -d_{i^{\prime}}^{j} \right) \cdot p_{i^{\prime}}}},\ \ p_i=\small{\frac{\exp \left( \mathcal{S} _i \right)}{\Sigma _{i^{\prime}}\exp \left( \mathcal{S} _{i^{\prime}} \right)}}, \forall i\in \mathbf{g}$
            \STATE $\hat{x}_j=\underset{i=1}{\overset{h\times w}{\Sigma}}\mathbf{g}_{x,i}\cdot \hat{w}_{i}^{j}, \ \ \ \  \hat{y}_j=\underset{i=1}{\overset{h\times w}{\Sigma}}\mathbf{g}_{y,i}\cdot \hat{w}_{i}^{j}$
            \STATE $\mathcal{P} \gets \mathcal{P} \cup \left( \hat{x}_j,\hat{y}_j \right)$
            \STATE if 
            $j\in \mathcal{I} _+$, \ $\mathrm{labels}_j=1$, if $j\in \mathcal{I} _-$, $\mathrm{labels}_j=0$
            
        \ENDFOR
        \RETURN $\mathcal{P}$, labels
    \end{algorithmic}
\end{algorithm}
\subsection{LLaVA Encoder}
\label{LLaVAEncoder}
To encode image and text features in parallel, we follow the approach in works \cite{lai2024lisa,wu2024see} and instantiate $\mathcal{G}_{\mathcal{T}}$ with LLaVA \cite{liu2024visual}. Specifically, $\mathcal{G}_{\mathcal{T}}$ consists of a CLIP \cite{radford2021learningclip} model for processing image features and a LLaMA \cite{touvron2023llama} model for processing text features. First, the CLIP model convolves $\mathbf{x}_{img}$ into $N_t$ image patches, each of which is then encoded by a series of stacked vision transformers. Next, these image embeddings are projected via MLPs to the same dimension as the text features and embedded into the text instructions $\mathbf{x}_{txt}$ before being fed into the LLaMA, a large language model. Note that the parameters of $\mathcal{G}_{\mathcal{T}}$  are frozen. We utilize LoRA ~\cite{iclr2022lora} for efficient fine-tuning as work~\cite{lai2024lisa}.

\subsection{SAM Mask Decoder}
\label{SAMMaskDecoder}
To decode the segmentation mask based on the text instructions $\mathbf{x}_{txt}$, we deploy SAM \cite{kirillov2023segment} model as the visual backbone \cite{wu2024see}. First, the SAM model uses a prompt encoder to project sparse embeddings, \wrt  
$\boldsymbol{h}_{seg}^{\left( l_k \right)}, \mathcal{P}$
as queries. Also, the SAM image encoder, composed of a series of vision transformers, encodes visual embeddings of $\mathbf{x}_{img}$ as keys. Next, queries and keys interact through $2$ layers of the TwoWayAttention Block before finally being decoded into the mask.

\subsection{Training Objectives}
\label{TrainingObjectives}
To infuse segmentation capabilities into the LMMs $\mathcal{G}_{\theta}$, we jointly optimize the text generation loss $\mathcal{L}_{txt}$ in $\mathcal{G}_{\mathcal{T}}$, and the segmentation mask loss $\mathcal{L}_{mask}$ in 
$\mathcal{G}_{\mathcal{V}}$  \cite{lai2024lisa,wu2024see}.
Specifically, we use cross-entropy loss for 
$\mathcal{L}_{txt}$, pixel-wise binary cross-entropy (BCE) loss and DICE loss for $\mathcal{L}_{mask}$ as 
\begin{align}
\begin{aligned}
\mathcal{L} _{txt}&=\mathcal{L} _{ce}(\hat{\mathbf{y}}_{txt},\mathbf{y}_{txt}),
\\
\mathcal{L} _{mask}=\lambda _{bce}\mathcal{L} _{bce}&(\hat{\mathbf{M}},\mathbf{M})+\lambda _{dice}\mathcal{L} _{dice}(\hat{\mathbf{M}},\mathbf{M}),
\label{eq:losses}
\end{aligned}
\end{align}
where $\lambda_{bce}$ and $\lambda_{dice}$ are the loss weights, $\mathbf{y}_{txt}$ and $\mathbf{M}$ are the ground-truth targets. The overall objective $\mathcal{L}$ aggregates those losses in Eq.~\eqref{eq:losses}, weighted by $\lambda_{txt}$ and $\lambda_{mask}$ as
\begin{equation}
    \mathcal{L} = \lambda_{txt} \mathcal{L}_{txt} + \lambda_{mask} \mathcal{L}_{mask}.
\end{equation}

\begin{table*}[htbp]
    \centering
    \vspace{-0.2cm}
\caption{Comparisons of the state-of-the-art reasoning segmentation results on ReasonSeg dataset. * means results are reproduced by the official model. `ft' denotes using 239 reasoning segmentation samples to fine-tune the model.
    \vspace{-0.2cm}
    }
    \resizebox{.75\linewidth}{!}
    {
    \label{table:reason_seg}   
    \tabcolsep=0.4cm
    {
        \begin{tabular}{l|cc|cc|cc|cc}
            \toprule
            
            \multirow{3}*{Method} & \multicolumn{2}{c|}{val} & \multicolumn{6}{c}{test} \\ 
            
            \specialrule{0em}{0pt}{1pt}
            \cline{2-9}
            \specialrule{0em}{0pt}{1pt}
            
            
            ~ & \multicolumn{2}{c|}{overall} & \multicolumn{2}{c|}{short query} & \multicolumn{2}{c|}{long query} & \multicolumn{2}{c}{overall} \\

            \specialrule{0em}{0pt}{1pt}
            \cline{2-9}
            \specialrule{0em}{0pt}{1pt}
            
            ~ & gIoU & cIoU & gIoU & cIoU & gIoU & cIoU & gIoU & cIoU \\ 
            
            \specialrule{0em}{0pt}{1pt}
            \hline
            \specialrule{0em}{0pt}{1pt}
            X-Decoder~\citep{zou2023generalized} & 22.6 & 17.9 & 20.4 & 11.6 & 22.2 & 17.5 & 21.7 & 16.3 \\

            Grounded-SAM~\citep{liu2023grounding} & 26.0 & 14.5 & 17.8 & 10.8 & 22.4 & 18.6 & 21.3 & 16.4 \\

            SEEM~\citep{zou2024seem} & 25.5 & 21.2 & 20.1 & 11.5 & 25.6 & 20.8 & 24.3 & 18.7 \\
            
            OVSeg~\citep{liang2023open} & 28.5 & 18.6 & 18.0 & 15.5 & 28.7 & 22.5 & 26.1 & 20.8  \\



            GRES~\citep{liu2023gres} & 22.4 & 19.9 & 17.6 & 15.0 & 22.6 & 23.8 & 21.3 & 22.0 \\    %
    
            \specialrule{0em}{0pt}{1pt}
            \hline
            \specialrule{0em}{0pt}{1pt}
            
            *SESAME~\cite{wu2024see} & {34.8} & {39.1} & {28.3} & {27.6} & {31.6} & {32.7}& {30.5} & {30.4} \\
            
            LLaVA1.5-7B + OVSeg~\cite{lai2024lisa} & 38.2 & 23.5 & 24.2 & 18.7 & 44.6 & 37.1 & 39.7 & 31.8 \\

            LISA-7B~\cite{lai2024lisa} & 52.9 & 54.0 & 40.6 & 40.6 & 49.4 & 51.0 & 47.3 & 48.4 \\
            
            
            
            
            
            
            LISA-7B-LLaVA1.5 (ft)~\cite{lai2024lisa} & \textbf{61.3} & 62.9 & {48.3} & {46.3} & 57.9 & 59.7 & 55.6 & 56.9 \\

            
            
            
                        
            \specialrule{0em}{0pt}{1pt}
            \hline
            \specialrule{0em}{0pt}{1pt}

            \toolname-7B-LLaVA1.5 (ft)  & 59.8 & \textbf{67.6} & \textbf{52.6} &\textbf{49.5}  & \textbf{60.4} & \textbf{61.0} &\textbf{58.5}  &\textbf{58.6} 
            \\
            \bottomrule            
        \end{tabular}
    }
    }
    \vspace{-0.5cm}
\end{table*}

\vspace{-0.2cm}
\section{Experiment}


\vspace{-0.1cm}
\subsection{Experimental Setting}
\label{exp:setting}
\vspace{-0.1cm}
\paragraph{Network Architecture.}
We employ LLaVA 1.5-7B~\cite{liu2024visual} as the base language model $\mathcal{G}_{\mathcal{T}}$ and the ViT-H SAM~\citep{kirillov2023segment} as the visual backbone $\mathcal{G}_{\mathcal{V}}$. For image encoding, we adopt the clip-vit-large-patch14-336, which takes 336$\times$336 pixel images as input. The projection layer $\gamma$ consists of a series of stacked MLPs with channel dimensions [512, 4096, 4096].

\vspace{-0.3cm}
\paragraph{Implementation Details.}
We train on 4 NVIDIA 24GB 3090 GPUs for 20 epochs around 24 hours. We deploy DeepSpeed~\citep{rasley2020deepspeed} engine for distributed training, with a batch size per device of 2 and, a gradient accumulation step of 10. The AdamW~\citep{loshchilov2017decoupled} optimizer is initialized with a learning rate of 0.0003 and no weight decay (set to 0). The learning rate is updated by the WarmupDecayLR scheduler, with 100 warmup iterations. 
The weights for $\mathcal{L} _{mask}$ and $\mathcal{L} _{txt}$ are set to 1.0, and the BCE loss weight $\lambda_{bce}$ is set to 2.0 and the DICE loss weight $\lambda_{dice}$ to 0.5 by default. Each image is randomly assigned up to 3 categories before being decorated with a question-and-answer template. 
\vspace{-0.4cm}
\paragraph{Datasets.} 
We follow prior work LISA \cite{lai2024lisa} to organize data structure, which typically consists of three types of datasets: (1) As for semantic segmentation dataset, we use ADE20K~\citep{zhou2017scene} and COCO-Stuff~\citep{caesar2018coco}, PACO-LVIS~\citep{ramanathan2023paco}, and PASCAL-Part~\citep{chen2014detect}; (2) As for referring segmentation dataset, we use refCLEF, refCOCO, refCOCO+~\citep{kazemzadeh2014referitgame}, refCOCOg~\citep{mao2016generation}, and ReasonSeg~\cite{lai2024lisa}; (3) As for the visual question answering (VQA) dataset, we use LLaVA-Instruct-150k for LLaVA v1.5~\citep{liu2024visual}. Also, to teach READ to overcome false premises, we include FP-RefCOCO(+/g) ~\cite{wu2024see}, R-RefCOCO~\cite{wu2024Rrefcoco} for false premises assessment.

\vspace{-0.4cm}
\paragraph{Evaluation Metrics.}
We adhere to the practices established in prior works~\citep{kazemzadeh2014referitgame, mao2016generation, lai2024lisa} by employing two evaluation metrics: gIoU and cIoU.
The gIoU metric is calculated as the mean of the Intersection-over-Union (IoU) values across all individual images, and cIoU is computed by the cumulative intersection across the cumulative union.

\subsection{Results on ReasonSeg Dataset}
\label{exp:reasonseg}
\paragraph{Comparison with the State-of-the-Art.} 
To evaluate the performance of READ on ReasonSeg dataset, 
we use \textit{train} set for training and validate the performance on \textit{val} set and \textit{test} set. Table~\ref{table:reason_seg} reveals that our \toolname model achieves significant performance gains in both gIoU and cIoU scores, particularly in the more challenging long query scenarios.
Specifically, 
the READ-7B outperforms LISA-7B-LLaVAv1.5 on ReasonSeg dataset, with a 4.7$\%$ higher cIoU on the val set and a 2.9$\%$ improvement gIoU in the overall test set. In Table~\ref{table:reason_seg_13B}, the READ-13B shows a 0.6$\%$ overall advantage, while in the short query subset, it has a 3.1$\%$ lead. 
This validates the effectiveness of our approach in handling complex reasoning tasks and generating accurate segmentation masks.
\begin{table}[htbp]
    \centering
    \vspace{-0.6cm}
\caption{Comparisons of the state-of-the-art on the ReasonSeg \textit{test} set using LLaVA 1.5-13B~\cite{liu2024improved} as the base language model.
    \vspace{-0.2cm}
    }
    \resizebox{1\linewidth}{!}
    {
    \label{table:reason_seg_13B}   
    \tabcolsep=0.15cm
    {
        \begin{tabular}{l|cc|cc|cc}
            \toprule
            
            \multirow{2}*{Method} & \multicolumn{2}{c|}{short query} & \multicolumn{2}{c|}{long query} & \multicolumn{2}{c}{overall} \\ 
            
            \specialrule{0em}{0pt}{1pt}
            \cline{2-7}
            \specialrule{0em}{0pt}{1pt}
            
            ~ & gIoU & cIoU & gIoU & cIoU & gIoU & cIoU \\ 
            
            \specialrule{0em}{0pt}{1pt}
            \hline
            \specialrule{0em}{0pt}{1pt}
 
            LISA-13B-LLaVA1.5 (ft) ~\cite{lai2024lisa}  & {55.4} &{50.6}  & {63.2} & \textbf{65.3} &{61.3}  &{62.2}\\

            \toolname-13B-LLaVA1.5 (ft)  &\textbf{55.4} &\textbf{53.7}  & \textbf{64.4} & {65.1} &\textbf{62.2}  &\textbf{62.8}            
            \\

            \bottomrule   
        \end{tabular}
    }
    }
    \vspace{-0.4cm}
\end{table}

\begin{figure*}[htbp]
\begin{center}
\vspace{-0.3cm}
\includegraphics[width=0.88\linewidth]{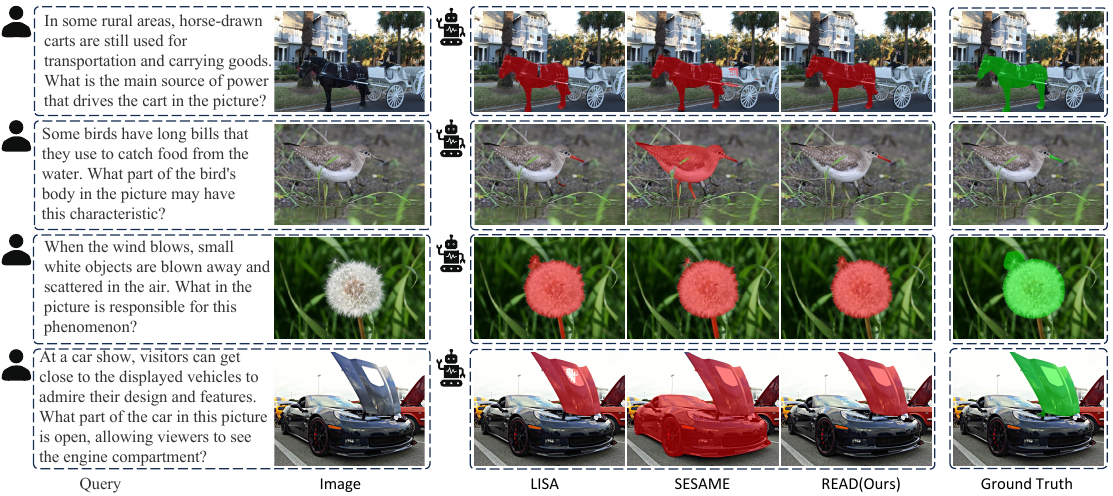}
\end{center}
\vspace{-0.7cm}
\caption{Visual comparison among \toolname (ours) and prior works on the ReasonSeg \textit{val} set. Refer to Appendix~\ref{sup:additional_qualitative_results} for more  illustrations.}
\label{fig:vis_comp}
\vspace{-0.6cm}
\end{figure*}


\subsection{Results on RefCOCO(+/g) Dataset}
\label{exp:referseg}
\paragraph{Comparison with the State-of-the-Art.} To demonstrate the efficacy of the proposed READ in the referring segmentation task, we conduct a comparative analysis with the existing state-of-the-art method, as detailed in Table~\ref{table:refer_seg}.
Our evaluation encompasses the refCOCO, refCOCO+, and refCOCOg \textit{val} set and \textit{test} set.
The outcomes reveal that our model outperforms existing methods across various referring segmentation tasks. Specifically, READ achieves 3.2$\%$ higher cIoU on refCOCO val and 2.4$\%$ higher on refCOCO+ val set. On the refCOCOg set, READ shows an advantage of 2.2$\%$ higher on val(U) and 0.8$\%$ higher on test(U).
Overall, READ consistently performs better or on par than LISA-7B across all subsets.
\subsection{Results on FP-RefCOCO(+/g) Dataset}
\label{exp:fpreferseg} 
\paragraph{Comparison with the State-of-the-Art.} To determine whether READ can overcome false premises, we assess its generation ability on augmented FP-RefCOCO(+/g) dataset. Given that LISA~\cite{lai2024lisa} is trained on positive samples only, in a fashion which always encourages the model to output a mask, even if the object described in the query does not actually exist. As a result, LISA suffers from catastrophic forgetting of previous skills after fine-tuning \cite{wu2024see}. We follow SESAME~\cite{wu2024see}, using see scores for the binary classification accuracy and cIoU for the segment capability. Table~\ref{table:refer_seg_see_segment} shows that READ surpasses SESAME ~\cite{wu2024see} in the ``see" task across all datasets. Specifically, READ improves by 3.03$\%$ on FP-RefCOCO, 3.51$\%$ on FP-RefCOCO+, and 2.89$\%$ on FP-RefCOCOg. For the ``segment" task, READ edges out SESAME, with a 3.57$\%$ advantage on FP-RefCOCO, 3.73$\%$ on FP-RefCOCO+, and 2.33$\%$ on FP-RefCOCOg. This indicates READ’s capability in detecting queried 
objects while generating segmentation masks even under false premises. 
 \vspace{-0.2cm}
\begin{table}[htbp]
    \centering    \caption{Comparisons of the state-of-the-art referring segmentation cIoU on RefCOCO(+/g) dataset.}
    \vspace{-0.2cm}
    \resizebox{0.95\linewidth}{!}
    {
    \label{table:refer_seg}   
    \vspace{-0.2cm}
    \tabcolsep=0.1cm
    {
        \begin{tabular}{l|ccc|ccc|cc}
            \toprule
            
            \multirow{2}*{Method} & \multicolumn{3}{c|}{refCOCO} & \multicolumn{3}{c|}{refCOCO+}  & \multicolumn{2}{c}{refCOCOg} \\ 
            
            \specialrule{0em}{0pt}{1pt}
            \cline{2-9}
            \specialrule{0em}{0pt}{1pt}
            
            ~ & val & testA & testB & val & testA & testB & val(U) & test(U) \\ 
                 
            \specialrule{0em}{0pt}{1pt}
            \hline
            \specialrule{0em}{0pt}{1pt}

            MCN~\citep{luo2020multi} & 62.4 & 64.2 & 59.7 & 50.6 & 55.0 & 44.7 & 49.2 & 49.4 \\

            VLT~\citep{ding2021vision} & 67.5 & 70.5 & 65.2 & 56.3 & 61.0 & 50.1 & 55.0 & 57.7 \\

            CRIS~\citep{wang2022cris} & 70.5 & 73.2 & 66.1 & 62.3 & 68.1 & 53.7 & 59.9 & 60.4 \\

            LAVT~\citep{yang2022lavt} & 72.7 & 75.8 & 68.8 & 62.1 & 68.4 & 55.1 & 61.2 & 62.1 \\
                        
            X-Decoder~\citep{zou2023generalized} & - & - & - & - & - & - & 64.6 & -  \\

            ReLA~\citep{liu2023gres} & 73.8 & 76.5 & 70.2 & {66.0} & {71.0} & 57.7 & 65.0 & 66.0 \\

            SEEM~\citep{zou2024seem} & - & - & - & - & - & - & 65.7 & -  \\
                        
            
            SESAME~\cite{wu2024see} & {74.7} & {-} & {-} & 64.9 & {-} & {-} & {66.1} & {-} \\

            LISA-7B~\cite{lai2024lisa} & {74.9} & {79.1} & {72.3} & 65.1 & 70.8 & {58.1} & {67.9} & {70.6} \\
            
            \specialrule{0em}{0pt}{1pt}
            \hline
            \specialrule{0em}{0pt}{1pt}

            READ & \textbf{78.1} &  \textbf{80.2}& \textbf{73.2} & \textbf{68.4} & \textbf{{73.7}} &\textbf{60.4} & \textbf{{70.1}} & \textbf{71.4} \\
            
            \bottomrule            
        \end{tabular}
    }
    }
\end{table}
\vspace{-0.3cm}
\subsection{Qualitative Results}
\label{exp:qualitative}
In Fig.~\ref{fig:vis_comp},
READ qualitatively outperforms prior works, such as LISA \cite{lai2024lisa} and SESAME \cite{wu2024see} in reasoning segmentation. Remarkably, READ is capable of handling fine-grained visual grounding tasks, as illustrated in the $2^{nd}$ row. 
\subsection{Ablation Study}
\label{exp:ablation}
In this section, we conduct an ablation study to analyze the contribution of each component. We report the gIoU and cIoU performance on the \textit{val} set of ReasonSeg dataset.

\begin{table}[htbp]
    \centering
    \caption{
    Comparisons of the state-of-the-art ``see'' and ``segment'' results on augmented FP-refcoco(+/g) \textit{val} set.
    Numbers of LISA, Cascading, and SESAME are cited from~\cite{wu2024see}.
    ``FP'' is the abbreviation for False Premise, which denotes a query for an object that is absent from the provided image.}
    \vspace{-0.3cm}
\resizebox{1\linewidth}{!}
    {
        \begin{tabular}{ l | c c | c c | c c }
            \toprule
            
            \multirow{2}*{Method} & \multicolumn{2}{c|}{FP-RefCOCO} & \multicolumn{2}{c|}{FP-RefCOCO+}  & \multicolumn{2}{c}{FP-RefCOCOg} \\ 
            
            \specialrule{0em}{0pt}{1pt}
            \cline{2-7}
            \specialrule{0em}{0pt}{1pt}
            ~ & See & Segment & See & Segment & See & Segment \\ 
            \specialrule{0em}{0pt}{1pt}
            \hline
            \specialrule{0em}{0pt}{1pt}
            LISA-7B \cite{lai2024lisa} & 51.36 & 44.00 & 51.32 &  39.62 & 51.25 & 39.64 \\
            Cascading \cite{wu2024see} & 75.59 &  55.18 & 75.03 & 48.64 & 76.07 & 49.98 \\
            SESAME \cite{wu2024see} & {79.84} & {57.93} & {80.00} & {50.81} & {81.78} & {53.79} \\
            \specialrule{0em}{0pt}{1pt}
            \hline
            \specialrule{0em}{0pt}{1pt}
            \toolname & \textbf{82.87} & \textbf{61.50} & \textbf{83.51} & \textbf{54.54} & \textbf{84.67} & \textbf{56.12}\\
            \bottomrule            
        \end{tabular}
    }
    \label{table:refer_seg_see_segment}  
    \vspace{-0.4cm}
\end{table}
\vspace{-0.4cm}
\paragraph{Effect of similarity as point.}
To assess each component of the SasP module, we refer to Table~\ref{table:ablation}, which shows that \(\mathcal{P}\)\textsubscript{prompt} improves cIoU by 7\% over \texttt{<SEG>}\textsubscript{prompt} alone, while \(\mathcal{P}\)\textsubscript{DtoC} provides an additional 3\% gain. 
\vspace{-0.4cm}
\begin{table}[htbp]
    \footnotesize
    \centering
    \caption{Ablation study on similarity as points (SasP).}
    \vspace{-0.36cm}
    \label{table:ablation}
    \tabcolsep=0.2cm
    \resizebox{0.7\linewidth}{!}
    {
    \begin{tabular}{c|ccc|cc}
        \toprule
        Exp. ID &  \texttt{<SEG>}\textsubscript{prompt} & $\mathcal{P}$\textsubscript{prompt}&$\mathcal{P}$\textsubscript{DtoC} & gIoU & cIoU \\
        \midrule
        1 & \Checkmark & & & 51.2 & 57.6 
        \\
        2 &  & \Checkmark& & 37.4 & 29.2 
        \\
        3 & \Checkmark & \Checkmark & & 56.4 & 64.6  \\
        4 & \Checkmark & \Checkmark & \Checkmark &\textbf{59.8} & \textbf{67.6} \\
        \bottomrule
    \end{tabular}
    }
\vspace{-0.2cm}
\end{table}

\paragraph{Effect of vision backbone.} Table~\ref{tab:backbone} reveals that larger vision backbones generally yield better performance. SAM-ViT-Huge achieves the highest cIoU (67.6\%) at 336px, indicating a trade-off between model size and input resolution.

\begin{table}[htbp]
  \centering
  \vspace{-0.15cm}
  \caption{Ablation study on vision backbone.}
    \vspace{-0.35cm}
\resizebox{0.7\linewidth}{!}{
    \begin{tabular}{l|l|cc|cc}
    \toprule
    \multirow{2}[4]{*}{Backbone} & \multirow{2}[4]{*}{Params} & \multicolumn{2}{c|}{CLIP@224px} & \multicolumn{2}{c}{CLIP@336px} \\
\cmidrule{3-6}          &       & gIoU  & cIoU  & gIoU  & cIoU \\
    \midrule
    SAM-ViT-Base &\raisebox{-0.5ex}{\textasciitilde}86M  & 54.6     & 57.8     & 55.6     & 61.9 \\
    SAM-ViT-Large & \raisebox{-0.5ex}{\textasciitilde}307M & 58.7     & 64.6     & \textbf{60.1}     & 65.2 \\
    SAM-ViT-Huge & \raisebox{-0.5ex}{\textasciitilde}636M & \textbf{59.9}     & \textbf{66.0}     & 59.8     & \textbf{67.6} \\
    \bottomrule
    \end{tabular}%
}
  \label{tab:backbone}%
  \vspace{-0.51cm}
\end{table}%

\section{Conclusion}
\vspace{-0.1cm}
In this paper, we investigate how the \texttt{<SEG>} token works, and we have found that what the \texttt{<SEG>} token contributes to is semantic similarity, akin to that of direct mentions in text, and that it aligns textual semantics with its visual space based on our findings. We therefore present READ, 
to guide LMMs where to ``attend'' when reasoning interactively by regarding similarity as points, which can be seamlessly applied to existing \texttt{<SEG>}-like paradigms with negligible effort and boosts performance remarkably. For future work, we aim to shed light on bridging textual vocabulary embeddings with the visual space to enhance multimodal alignment.

\vspace{-0.3cm}
\section*{Acknowledgments}
\vspace{-0.2cm}
\fontsize{9.3pt}{10.6pt}\selectfont This work was supported by Dejing Dou's Research Startup Fund from Fudan University and the computations in this research were performed using the CFFF platform of Fudan University.

\normalsize

{
    \small
    \bibliographystyle{ieeenat_fullname}
    \bibliography{main}

@String(CVPR= {IEEE Conf. Comput. Vis. Pattern Recog.})

@String(ICCV= {Int. Conf. Comput. Vis.})

@String(ECCV= {Eur. Conf. Comput. Vis.})

@String(TIP  = {IEEE Trans. Image Process.})

@String(ICLR = {Int. Conf. Learn. Represent.})

@String(CVPR  = {CVPR})

@String(ICCV  = {ICCV})

@String(ECCV  = {ECCV})

@String(TIP   = {IEEE TIP})

@String(ICLR  = {ICLR})

@inproceedings{wu2024Rrefcoco,
  title={Towards Robust Referring Image Segmentation},
  author={Wu, Jianzong and Li, Xiangtai and Li, Xia and Ding, Henghui and Tong, Yunhai and Tao, Dacheng},
  booktitle={TIP},
  year={2024},
}

@inproceedings{lai2024lisa,
  title={{LISA}: Reasoning segmentation via large language model},
  author={Lai, Xin and Tian, Zhuotao and Chen, Yukang and Li, Yanwei and Yuan, Yuhui and Liu, Shu and Jia, Jiaya},
  booktitle={CVPR},
  year={2024}
}

@inproceedings{hu2016segmentation,
  title={Segmentation from natural language expressions},
  author={Hu, Ronghang and Rohrbach, Marcus and Darrell, Trevor},
  booktitle={ECCV},
  year={2016}
}

@inproceedings{wu2024see,
  title={{See Say and Segment: Teaching LMMs to Overcome False Premises}},
  author={Wu, Tsung-Han and Biamby, Giscard and Chan, David and Dunlap, Lisa and Gupta, Ritwik and Wang, Xudong and Gonzalez, Joseph E and Darrell, Trevor},
  booktitle={CVPR},
  year={2024}
}

@inproceedings{rasheed2024glamm,
  title={{GlaMM}: Pixel grounding large multimodal model},
  author={Rasheed, Hanoona and Maaz, Muhammad and Shaji, Sahal and Shaker, Abdelrahman and Khan, Salman and Cholakkal, Hisham and Anwer, Rao M and Xing, Eric and Yang, Ming-Hsuan and Khan, Fahad S},
  booktitle={CVPR},
  year={2024}
}

@inproceedings{ren2024pixellm,
  title={{PixelLM}: Pixel reasoning with large multimodal model},
  author={Ren, Zhongwei and Huang, Zhicheng and Wei, Yunchao and Zhao, Yao and Fu, Dongmei and Feng, Jiashi and Jin, Xiaojie},
  booktitle={CVPR},
  year={2024}
}

@inproceedings{xia2024gsva,
  title={{GSVA}: Generalized segmentation via multimodal large language models},
  author={Xia, Zhuofan and Han, Dongchen and Han, Yizeng and Pan, Xuran and Song, Shiji and Huang, Gao},
  booktitle={CVPR},
  year={2024}
}

@inproceedings{liu2024visual,
  title={Visual instruction tuning},
  author={Liu, Haotian and Li, Chunyuan and Wu, Qingyang and Lee, Yong Jae},
  booktitle={NeurIPS},
  year={2024}
}

@inproceedings{long2015fcn,
  title={Fully convolutional networks for semantic segmentation},
  author={Long, Jonathan and Shelhamer, Evan and Darrell, Trevor},
  booktitle={CVPR},
  year={2015}
}

@inproceedings{ronneberger2015unet,
  title={{U-Net}: Convolutional networks for biomedical image segmentation},
  author={Ronneberger, Olaf and Fischer, Philipp and Brox, Thomas},
  booktitle={MICCAI},
  year={2015},
}

@inproceedings{badrinarayanan2017segnet,
  title={{SegNet}: A deep convolutional encoder-decoder architecture for image segmentation},
  author={Badrinarayanan, Vijay and Kendall, Alex and Cipolla, Roberto},
  booktitle={TPAMI},
  year={2017},
}

@inproceedings{he2017mask,
  title={{Mask R-CNN}},
  author={He, Kaiming and Gkioxari, Georgia and Doll{\'a}r, Piotr and Girshick, Ross},
  booktitle={CVPR},
  year={2017}
}

@inproceedings{liu2018path,
  title={Path aggregation network for instance segmentation},
  author={Liu, Shu and Qi, Lu and Qin, Haifang and Shi, Jianping and Jia, Jiaya},
  booktitle={CVPR},
  year={2018}
}

@inproceedings{carion2020DETR,
  title={End-to-end object detection with transformers},
  author={Carion, Nicolas and Massa, Francisco and Synnaeve, Gabriel and Usunier, Nicolas and Kirillov, Alexander and Zagoruyko, Sergey},
  booktitle={ECCV},
  year={2020},
}

@inproceedings{xiong2019upsnet,
  title={{UPSNet}: A unified panoptic segmentation network},
  author={Xiong, Yuwen and Liao, Renjie and Zhao, Hengshuang and Hu, Rui and Bai, Min and Yumer, Ersin and Urtasun, Raquel},
  booktitle={CVPR},
  year={2019}
}

@inproceedings{kirillov2019panopticps,
  title={Panoptic segmentation},
  author={Kirillov, Alexander and He, Kaiming and Girshick, Ross and Rother, Carsten and Doll{\'a}r, Piotr},
  booktitle={CVPR},
  year={2019}
}

@inproceedings{kirillov2019panopticfpn,
  title={Panoptic feature pyramid networks},
  author={Kirillov, Alexander and Girshick, Ross and He, Kaiming and Doll{\'a}r, Piotr},
  booktitle={CVPR},
  year={2019}
}

@inproceedings{ren2016faster,
  title={{Faster R-CNN}: Towards real-time object detection with region proposal networks},
  author={Ren, Shaoqing and He, Kaiming and Girshick, Ross and Sun, Jian},
  booktitle={TPAMI},
  year={2016},
}

@inproceedings{zou2024seem,
  title={Segment everything everywhere all at once},
  author={Zou, Xueyan and Yang, Jianwei and Zhang, Hao and Li, Feng and Li, Linjie and Wang, Jianfeng and Wang, Lijuan and Gao, Jianfeng and others},
  booktitle={NeurIPS},
  year={2024}
}

@inproceedings{iclr2022lora,
  author       = {Edward J. Hu and
                  Yelong Shen and
                  Phillip Wallis and
                  Zeyuan Allen{-}Zhu and
                  Yuanzhi Li and
                  Shean Wang and
                  Lu Wang and
                  Weizhu Chen},
  title        = {{LoRA}: Low-Rank Adaptation of Large Language Models},
  booktitle    = { {ICLR}},
  year         = {2022},
}

@inproceedings{radford2021learningclip,
  title={Learning transferable visual models from natural language supervision},
  author={Radford, Alec and Kim, Jong Wook and Hallacy, Chris and Ramesh, Aditya and Goh, Gabriel and Agarwal, Sandhini and Sastry, Girish and Askell, Amanda and Mishkin, Pamela and Clark, Jack and others},
  booktitle={ICML},
  year={2021},
}

@inproceedings{vaswani2017crossattention,
  title={Attention is all you need},
  author={Vaswani, Ashish and Shazeer, Noam and Parmar, Niki and Uszkoreit, Jakob and Jones, Llion and Gomez, Aidan N and Kaiser, {\L}ukasz and Polosukhin, Illia},
  booktitle={NeurIPS},
  year={2017}
}

@inproceedings{detgpt,
  author       = {Renjie Pi and
                  Jiahui Gao and
                  Shizhe Diao and
                  Rui Pan and
                  Hanze Dong and
                  Jipeng Zhang and
                  Lewei Yao and
                  Jianhua Han and
                  Hang Xu and
                  Lingpeng Kong and
                  Tong Zhang},
  title        = {{DetGPT}: Detect What You Need via Reasoning},
  booktitle      = {EMNLP},
  year         = {2023}
}

@inproceedings{wang2023visionllm,
  title={{VisionLLM}: large language model is also an open-ended decoder for vision-centric tasks},
  author={Wang, Wenhai and Chen, Zhe and Chen, Xiaokang and Wu, Jiannan and Zhu, Xizhou and Zeng, Gang and Luo, Ping and Lu, Tong and Zhou, Jie and Qiao, Yu and others},
  booktitle={NeurIPS},
  year={2023}
}

@inproceedings{mao2016generation,
  title={Generation and Comprehension of Unambiguous Object Descriptions},
  author={Mao, Junhua and Huang, Jonathan and Toshev, Alexander and Camburu, Oana and Yuille, Alan L and Murphy, Kevin},
  booktitle={CVPR},
  year={2016}
}

@inproceedings{kirillov2023segment,
  title={Segment anything},
  author={Kirillov, Alexander and Mintun, Eric and Ravi, Nikhila and Mao, Hanzi and Rolland, Chloe and Gustafson, Laura and Xiao, Tete and Whitehead, Spencer and Berg, Alexander C and Lo, Wan-Yen and others},
  booktitle={ICCV},
  year={2023}
}

@inproceedings{alayrac2022flamingo,
  title={Flamingo: a visual language model for few-shot learning},
  author={Alayrac, Jean-Baptiste and Donahue, Jeff and Luc, Pauline and Miech, Antoine and Barr, Iain and Hasson, Yana and Lenc, Karel and Mensch, Arthur and Millican, Katherine and Reynolds, Malcolm and others},
  booktitle={NeurIPS},
  year={2022}
}

@inproceedings{li2023blip,
  title={{BLIP-2}: Bootstrapping language-image pre-training with frozen image encoders and large language models},
  author={Li, Junnan and Li, Dongxu and Savarese, Silvio and Hoi, Steven},
  booktitle={ICML},
  year={2023}
}

@inproceedings{li2023otter,
  title={Otter: A multi-modal model with in-context instruction tuning},
  author={Li, Bo and Zhang, Yuanhan and Chen, Liangyu and Wang, Jinghao and Yang, Jingkang and Liu, Ziwei},
  booktitle={TPAMI},
  year={2025}
}

@inproceedings{rasley2020deepspeed,
  title={{DeepSpeed}: System Optimizations Enable Training Deep Learning Models with Over 100 Billion Parameters},
  author={Rasley, Jeff and Rajbhandari, Samyam and Ruwase, Olatunji and He, Yuxiong},
  booktitle={SIGKDD},
  year={2020}
}

@inproceedings{loshchilov2017decoupled,
  title={Decoupled weight decay regularization},
  author={Loshchilov, Ilya and Hutter, Frank},
  booktitle={ICLR},
  year={2017}
}

@inproceedings{zhou2017scene,
  title={Scene Parsing through {ADE20K} Dataset},
  author={Zhou, Bolei and Zhao, Hang and Puig, Xavier and Fidler, Sanja and Barriuso, Adela and Torralba, Antonio},
  booktitle={CVPR},
  year={2017}
}

@inproceedings{caesar2018coco,
  title={Coco-stuff: Thing and stuff classes in context},
  author={Caesar, Holger and Uijlings, Jasper and Ferrari, Vittorio},
  booktitle={CVPR},
  year={2018}
}

@inproceedings{ramanathan2023paco,
  title={{PACO}: Parts and Attributes of Common Objects},
  author={Ramanathan, Vignesh and Kalia, Anmol and Petrovic, Vladan and Wen, Yi and Zheng, Baixue and Guo, Baishan and Wang, Rui and Marquez, Aaron and Kovvuri, Rama and Kadian, Abhishek and others},
  booktitle={CVPR},
  year={2023}
}

@inproceedings{chen2014detect,
  title={Detect What You Can: Detecting and Representing Objects using Holistic Models and Body Parts},
  author={Chen, Xianjie and Mottaghi, Roozbeh and Liu, Xiaobai and Fidler, Sanja and Urtasun, Raquel and Yuille, Alan},
  booktitle={CVPR},
  year={2014}
}

@inproceedings{kazemzadeh2014referitgame,
  title={{ReferItGame}: Referring to Objects in Photographs of Natural Scenes},
  author={Kazemzadeh, Sahar and Ordonez, Vicente and Matten, Mark and Berg, Tamara},
  booktitle={EMNLP},
  year={2014}
}

@inproceedings{zou2023generalized,
  title={Generalized decoding for pixel, image, and language},
  author={Zou, Xueyan and Dou, Zi-Yi and Yang, Jianwei and Gan, Zhe and Li, Linjie and Li, Chunyuan and Dai, Xiyang and Behl, Harkirat and Wang, Jianfeng and Yuan, Lu and others},
  booktitle={CVPR},
  year={2023}
}

@inproceedings{liu2023gres,
  title={{GRES}: Generalized Referring Expression Segmentation},
  author={Liu, Chang and Ding, Henghui and Jiang, Xudong},
  booktitle={CVPR},
  year={2023}
}

@inproceedings{liang2023open,
  title={Open-Vocabulary Semantic Segmentation with Mask-adapted {CLIP}},
  author={Liang, Feng and Wu, Bichen and Dai, Xiaoliang and Li, Kunpeng and Zhao, Yinan and Zhang, Hang and Zhang, Peizhao and Vajda, Peter and Marculescu, Diana},
  booktitle={CVPR},
  year={2023}
}

@inproceedings{zhu2023minigpt,
  title={{MiniGPT-4}: Enhancing vision-language understanding with advanced large language models},
  author={Zhu, Deyao and Chen, Jun and Shen, Xiaoqian and Li, Xiang and Elhoseiny, Mohamed},
  booktitle={ICLR},
  year={2024}
}

@inproceedings{ye2024mplug,
  title={{mPLUG-Owl2}: Revolutionizing multi-modal large language model with modality collaboration},
  author={Ye, Qinghao and Xu, Haiyang and Ye, Jiabo and Yan, Ming and Hu, Anwen and Liu, Haowei and Qian, Qi and Zhang, Ji and Huang, Fei},
  booktitle={CVPR},
  year={2024}
}

@inproceedings{peng2023kosmos,
  title={Kosmos-2: Grounding Multimodal Large Language Models to the World},
  author={Peng, Zhiliang and Wang, Wenhui and Dong, Li and Hao, Yaru and Huang, Shaohan and Ma, Shuming and Wei, Furu},
  booktitle={ICLR},
  year={2024}
}

@inproceedings{zhang2023gpt4roi,
  title={{GPT4RoI}: Instruction Tuning Large Language Model on Region-of-Interest},
  author={Zhang, Shilong and Sun, Peize and Chen, Shoufa and Xiao, Min and Shao, Wenqi and Zhang, Wenwei and Chen, Kai and Luo, Ping},
  booktitle={ECCVW},
  year={2025}
}

@inproceedings{yang2022lavt,
  title={{LAVT}: Language-Aware Vision Transformer for Referring Image Segmentation},
  author={Yang, Zhao and Wang, Jiaqi and Tang, Yansong and Chen, Kai and Zhao, Hengshuang and Torr, Philip HS},
  booktitle={CVPR},
  year={2022}
}

@inproceedings{wang2022cris,
  title={{CRIS}: {CLIP-Driven} Referring Image Segmentation},
  author={Wang, Zhaoqing and Lu, Yu and Li, Qiang and Tao, Xunqiang and Guo, Yandong and Gong, Mingming and Liu, Tongliang},
  booktitle={CVPR},
  year={2022}
}

@inproceedings{ding2021vision,
  title={Vision-Language Transformer and Query Generation for Referring Segmentation},
  author={Ding, Henghui and Liu, Chang and Wang, Suchen and Jiang, Xudong},
  booktitle={ICCV},
  year={2021}
}

@inproceedings{luo2020multi,
  title={Multi-task Collaborative Network for Joint Referring Expression Comprehension and Segmentation},
  author={Luo, Gen and Zhou, Yiyi and Sun, Xiaoshuai and Cao, Liujuan and Wu, Chenglin and Deng, Cheng and Ji, Rongrong},
  booktitle={CVPR},
  year={2020}
}

@inproceedings{liu2023grounding,
  title={{Grounding DINO}: Marrying dino with grounded pre-training for open-set object detection},
  author={Liu, Shilong and Zeng, Zhaoyang and Ren, Tianhe and Li, Feng and Zhang, Hao and Yang, Jie and Li, Chunyuan and Yang, Jianwei and Su, Hang and others},
  booktitle={ECCV},
  year={2024}
}

@inproceedings{liu2024improved,
  title={Improved Baselines with Visual Instruction Tuning},
  author={Liu, Haotian and Li, Chunyuan and Li, Yuheng and Lee, Yong Jae},
  booktitle={CVPR},
  year={2024}
}

@inproceedings{touvron2023llama,
  title={{LLaMA}: Open and efficient foundation language models},
  author={Touvron, Hugo and Lavril, Thibaut and Izacard, Gautier and Martinet, Xavier and Lachaux, Marie-Anne and Lacroix, Timoth{\'e}e and Rozi{\`e}re, Baptiste and Goyal, Naman and Hambro, Eric and Azhar, Faisal and others},
  booktitle={arXiv:2302.13971},
  year={2023}
}

@inproceedings{yao2024mmca,
  title={Visual Grounding with Multi-modal Conditional Adaptation},
  author={Yao, Ruilin and Xiong, Shengwu and Zhao, Yichen and Rong, Yi},
  booktitle={ACM MM},
  year={2024}
}

@inproceedings{chen2022MDGT,
  title={Multi-modal dynamic graph transformer for visual grounding},
  author={Chen, Sijia and Li, Baochun},
  booktitle={CVPR},
  year={2022}
}
}
\clearpage
\setcounter{page}{1}
\maketitlesupplementary
\appendix

\noindent We provide supplementary material related to the main paper, arranged as follows:
\begin{enumerate}
    \item Additional implementation details (Appendix~\ref{sup:implementation_details})
    \item Additional Analysis (Appendix~\ref{sup:additional_analysis})
    \item Additional ablation study (Appendix~\ref{sup:additional_ablation_study})
    \item Additional qualitative results (Appendix~\ref{sup:additional_qualitative_results})
    \item Discussion (Appendix~\ref{sup:discussion})
\end{enumerate}

\vspace{-0.2cm}
\section{Additional Implementation Details}
\label{sup:implementation_details}
\subsection{Grid Search for Optimal Threshold}
\label{sup:GridSearchforOptimal Threshold}
Given that the threshold for the foreground mask has a significant impact on the IoU, to eliminate the bias introduced by manually setting the threshold (\eg, 0.5), we perform a grid search over the similarity map for each image with a step size of 0.01 to identify the optimal foreground mask. For each threshold $t$, we convert the similarity map into a binary mask by applying 
\vspace{-0.1cm}
\begin{align}
\begin{aligned}
\hat{\mathrm{M}}\left( x,y \right) =\begin{cases}
	1 \ \ if\,\,\mathcal{S} \left( x,y \right) \ge t\\
	0 \ \ if\,\,\mathcal{S} \left( x,y \right) <t\\
\end{cases},
\end{aligned}
\end{align}
where $\mathcal{S} \left( x,y \right)$ is the similarity score for each pixel at position $\left( x,y \right)$, $\hat{\mathrm{M}}\left( x,y \right)$ is the binary mask at that pixel. We calculate cIoU for all threshold values in the grid, and choose the threshold $t^{\prime}$ that maximizes the cIoU for the image 
\begin{align}
\begin{aligned}
t^{\prime}=\underset{t}{\mathrm{arg}\max}\left( \mathrm{cIoU}\left( t \right) \right).
\end{aligned}
\end{align}
Once the optimal threshold is selected for each image, we use it to generate the final binary masks for evaluation, which ensures that the comparison is fair and threshold-invariant.

\subsection{Model Architecture and Training}
\label{sup:app_training}
As for reasoning segmentation, we trained two models: READ-7B and READ-13B. For READ-7B, we initialize the parameters using the released SESAME model [42] to accelerate training, with the training dataset allocated in a 10:1:1:1:1:10 ratio. We employ LoRA~\cite{iclr2022lora} for efficient fine-tuning, using \( lora\_r = 8 \), and conduct end-to-end joint training. For READ-13B, we train it from scratch, using LLaVA 1.5-13B as the base model. Initially, we train it on the full dataset in a 10:10:2:3:1:1 ratio for about 8 epochs, and then fine-tune it with a ratio of 3:10:2:3:1:10, using a learning rate of 0.0001 and \( lora\_r = 64 \). As for referring segmentation, we maintain the same settings as those used for READ-7B in reasoning segmentation. All our code will be publicly available at \href{https://github.com/rui-qian/READ}{https://github.com/rui-qian/READ}.

\section{Additional Analysis}
\label{sup:additional_analysis}
\begin{figure*}[!th]
  \centering
\includegraphics[width=1\linewidth]{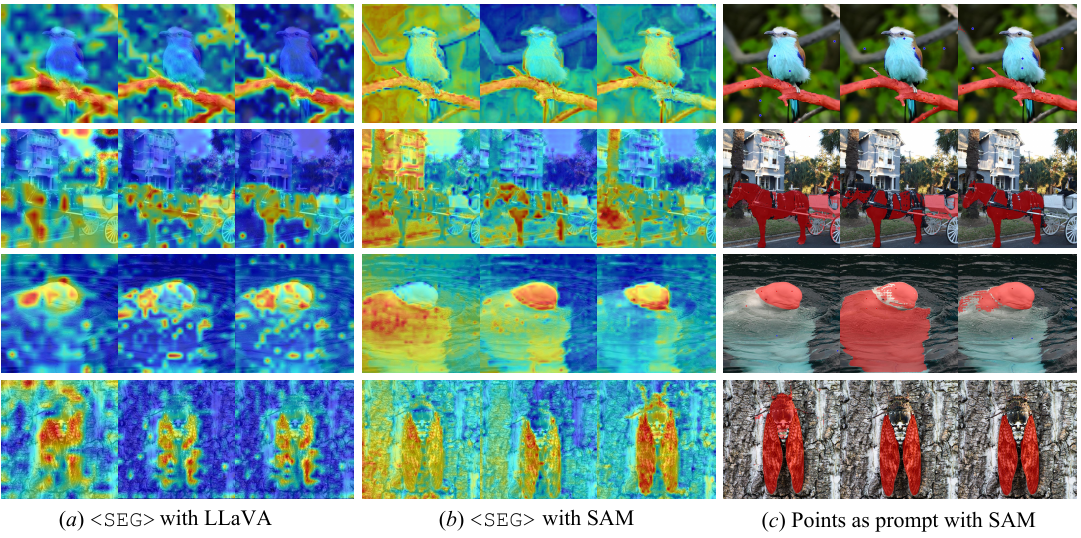}
\vspace{-0.7cm}
    \caption{
    Qualitative analysis of the 
  \texttt{<SEG>} token on the ReasonSeg \textit{val} set.  The $1^{st}$, $2^{nd}$, and $3^{rd}$ columns of $(a)$, $(b)$, and $(c)$ are LISA, SESAME, and READ (Ours) for comparisons, respectively. Points derived from $(a)$ serve as prompts with original SAM in $(c)$.}
\label{fig:Appendix01}
\end{figure*}

\begin{figure*}[!th]
  \centering
\includegraphics[width=0.95\linewidth]{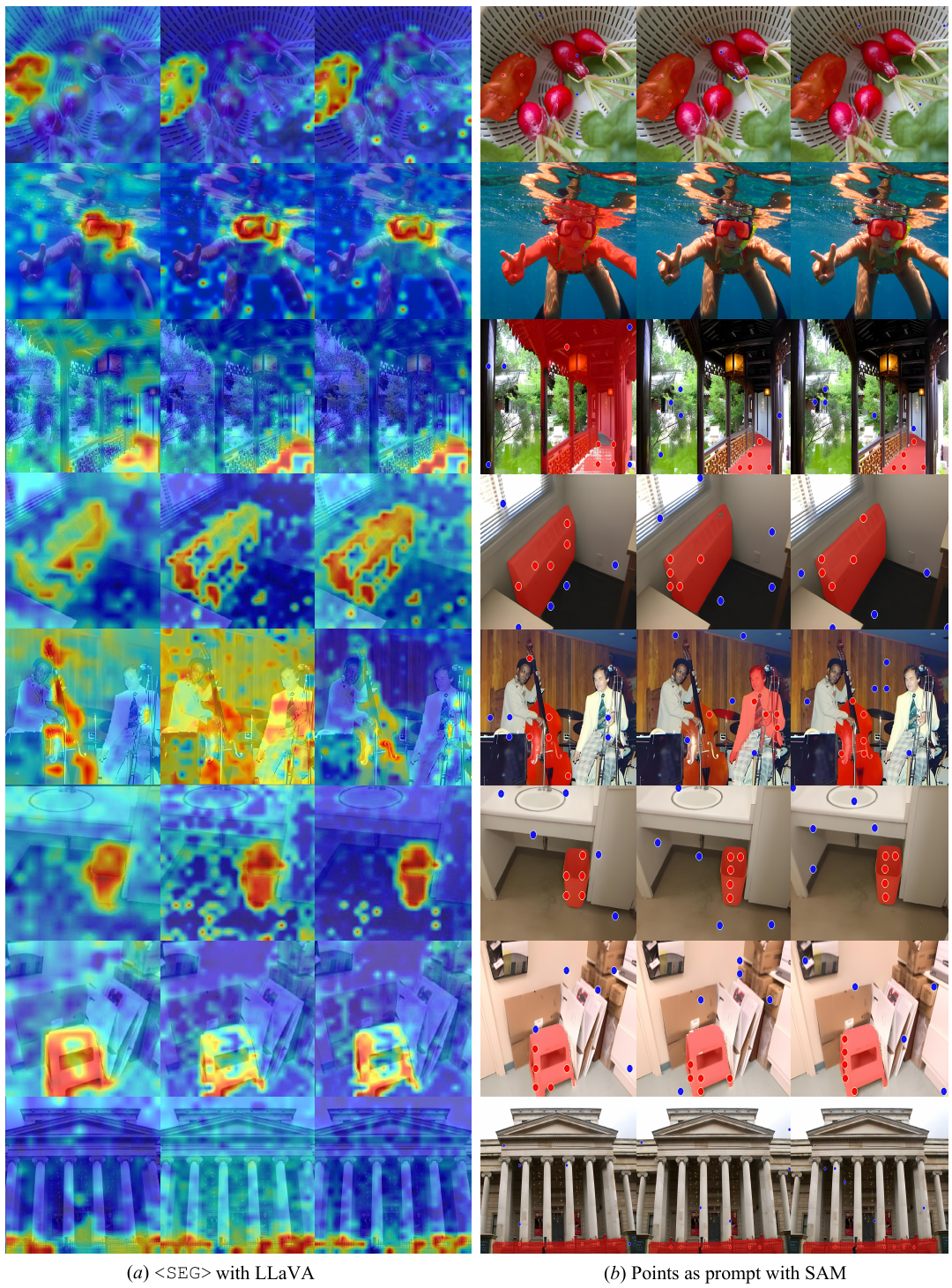}
\vspace{-0.3cm}
    \caption{
    Qualitative analysis of $\mathcal{P}$\textsubscript{prompt} (points as prompt) on the ReasonSeg \textit{val} set. The $1^{st}$, $2^{nd}$, and $3^{rd}$ columns of $(a)$, $(b)$ are LISA, SESAME, and READ (Ours) for comparisons, respectively. Points derived from $(a)$ serve as prompts with original SAM in $(b)$.}
\label{fig:Appendix02}
\end{figure*}

$(1)$ Fig.~\ref{fig:Appendix01} shows qualitative analysis of the 
  \texttt{<SEG>} token on the ReasonSeg \textit{val} set. Points derived from $(a)$ serve as prompts with original SAM in $(c)$. Similarity between the \texttt{<SEG>} token and image token embeddings stemming from the last hidden layer is computed by Eq.\eqref{eq:similarity}, \wrt LLaVA encoder in $(a)$ and SAM decoder in $(b)$. The consistency in $(a)$, $(b)$  indicates that the \texttt{<SEG>} token in
LMMs learns semantics similar to direct 
mentions in text, as observed in CLIP~\cite{radford2021learningclip}. Note that $1^{st}$ column in $(b)$ shows failure cases, indicating the existence of misalignment between the LLaVA encoder in $(a)$ and SAM decoder in $(b)$. Such observation sheds light on the interpretability of semantic alignment issues, where the LLaVA encoder generates accurate textual responses even in scenarios where the SAM decoder fails at segmentation, when eliciting  LISA~\cite{lai2024lisa} for reasoning explanations. In future work, we aim to further investigate the underlying connections behind this phenomenon. \ $(2)$
Fig.~\ref{fig:Appendix02} shows a qualitative analysis of $\mathcal{P}$\textsubscript{prompt} on the ReasonSeg \textit{val} set. 
   We first select several points with the highest similarity scores as positives (\textcolor{red}{red} in $(b)$) and an equal number of points with the lowest similarity scores as negatives (\textcolor{blue}{blue} in $(b)$). These points are then directly used as prompts instead of the \texttt{<SEG>} token, and are input into the original SAM model to generate the segmentation mask.
Columns in $(b)$ demonstrate that only relying on the selected similarity points as prompt can still generate a segmentation mask potentially.
\vspace{-0.1cm}
\section{Additional Ablation Study}
\label{sup:additional_ablation_study}
\paragraph{Effect of points ratios.} 
To explore how the ratios of positive, negative, and neutral points impact the performance of READ, we vary the positive and negative thresholds (\(t_{pos}\) and \(t_{neg}\)) as well as the number of points \(\mathcal{|P|}\). As the positive sample ratio (\(t_{pos}\)) increases, model performance improves, particularly when fewer points are used (\(\mathcal{|P|}\)=10). Also, increasing the number of points generally enhances performance, with the most significant improvements observed at \(\mathcal{|P|}\)=60, regardless of the \(t_{pos}\) setting.


\vspace{-0.2cm}
\begin{table}[htbp]
  \centering
  \caption{Ablation study on points ratios.}
  \vspace{-0.3cm}
      \tabcolsep=0.2cm
    \resizebox{0.65\linewidth}{!}
    {
    \begin{tabular}{cc|cc|cc|cc}
    \toprule
    \multirow{2}[4]{*}{$t_{pos}$} & \multirow{2}[4]{*}{$t_{neg}$} & \multicolumn{2}{c|}{$\mathcal{|P|}$=10} & \multicolumn{2}{c|}{$\mathcal{|P|}$=30} & \multicolumn{2}{c}{$\mathcal{|P|}$=60} \\
\cmidrule{3-8}          &       & gIoU  & cIoU  & gIoU  & cIoU  & gIoU  & cIoU \\
    \midrule
    0.8   & 0.2   & \textbf{58.94} & \textbf{65.16} & \textbf{59.75} & \textbf{67.62} & \textbf{59.71} & \textbf{68.17} \\
    0.7   & 0.3   & 58.48 & 64.00 & 58.82 & 65.32 & 59.20 & 67.70 \\
    0.6   & 0.4   & 58.59 & 64.27 & 58.66 & 65.00 & 58.93 & 66.69 \\
    \bottomrule
    \end{tabular}%
    }
 \label{tab:pointsportion}
\end{table}%
\vspace{-0.5cm}
\section{Additional Qualitative Results}
\vspace{-0.1cm}
\label{sup:additional_qualitative_results}

\begin{figure*}[!th]
  \centering
\includegraphics[width=0.9\linewidth]{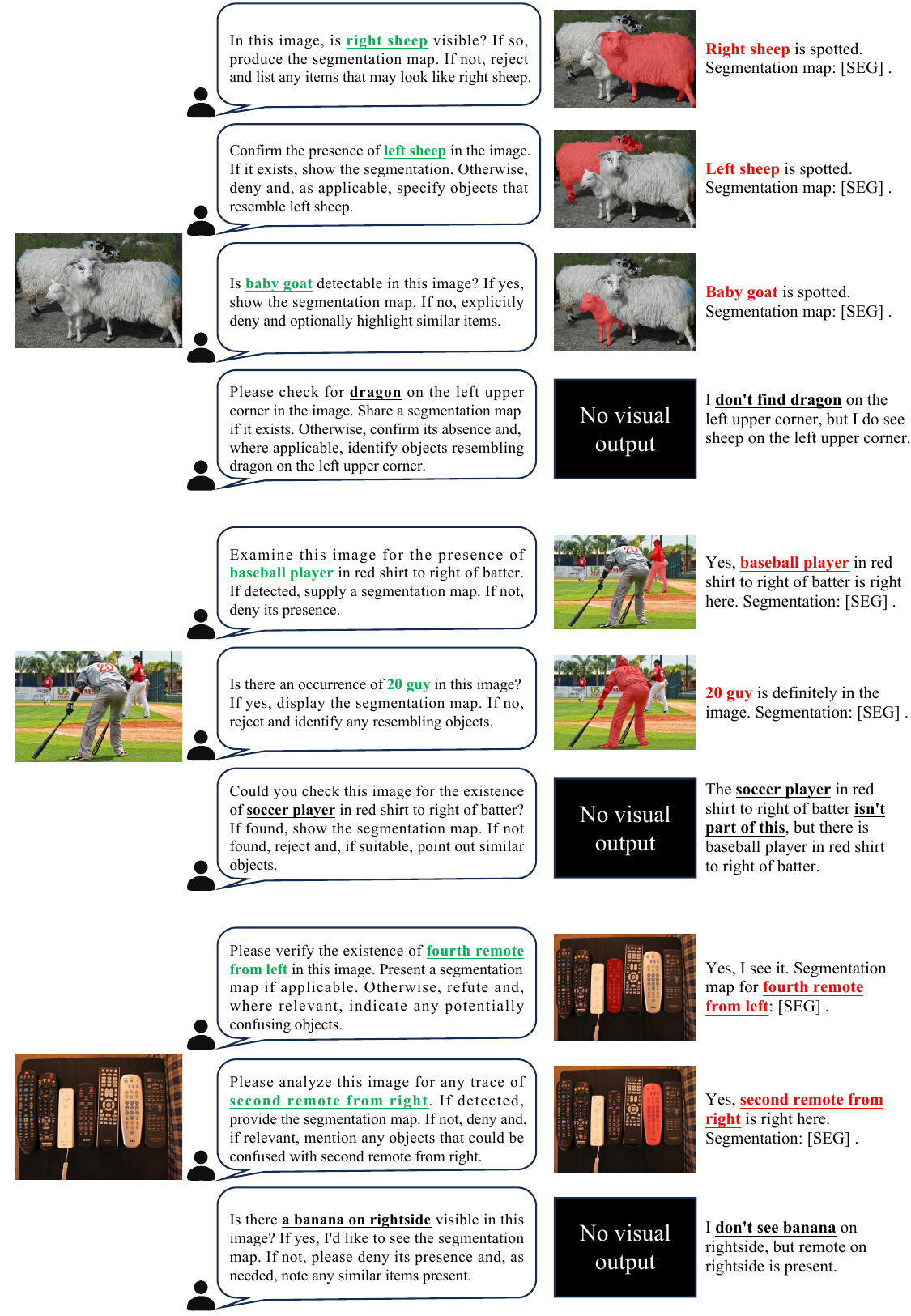}
\vspace{-0.3cm}
    \caption{Visualization on the FP-RefCOCO(+/g) \textit{val} set.
    }
    \label{fig:Appendixfprefcoco}
\end{figure*}

Fig.~\ref{fig:Appendixfprefcoco} shows qualitative results on the FP-RefCOCO(+/g) \textit{val} set. Also, READ retains the conversational ability of LLMs  while performing segmentation tasks and can refuse to output a mask when the queried object doesn't exist.

\begin{figure*}[!th]
  \centering
\includegraphics[width=0.9\linewidth]{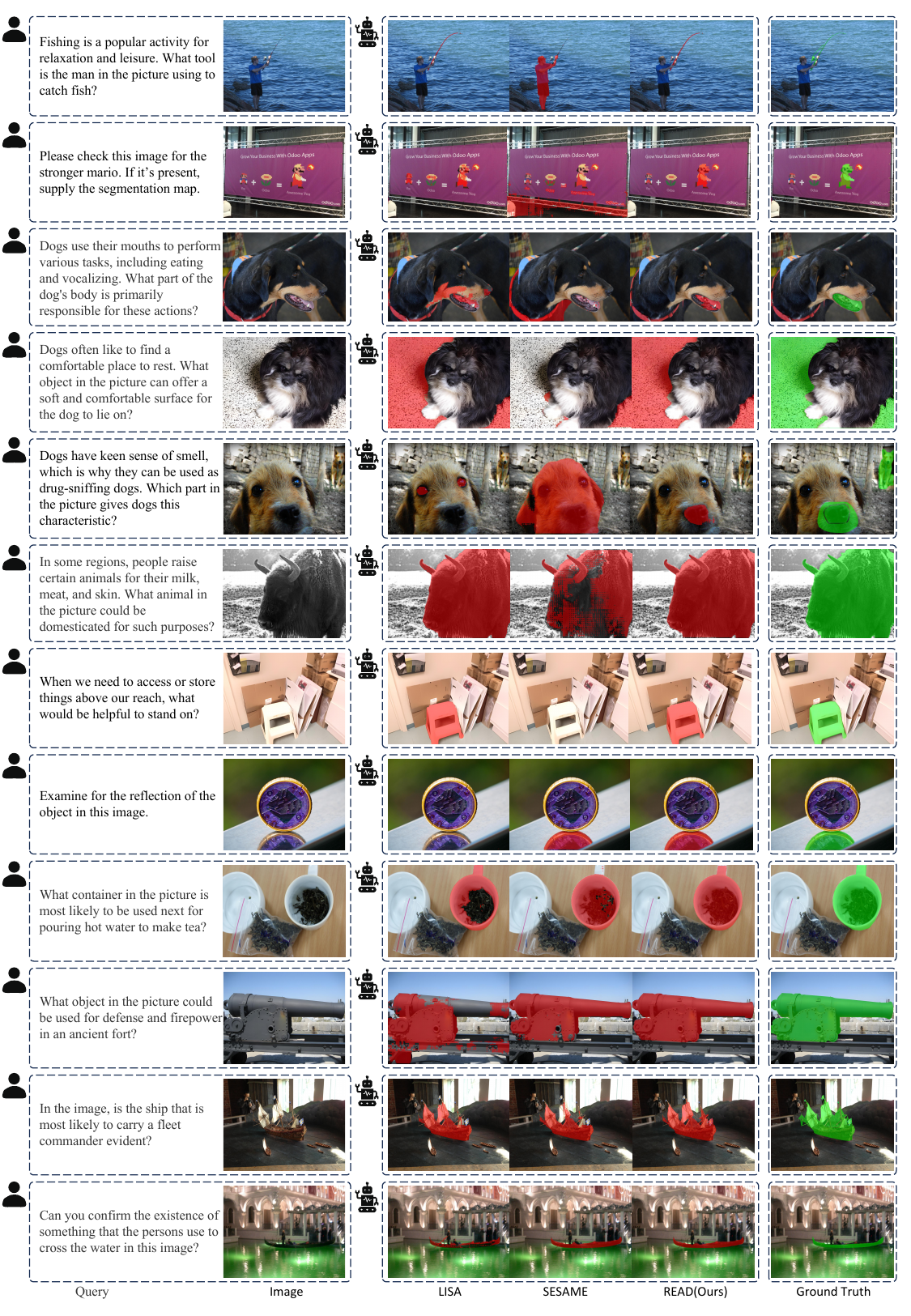}
\vspace{-0.3cm}
    \caption{Visual comparison among \toolname (ours) and prior works on the ReasonSeg \textit{val} set. 
    }
    \label{fig:Appendix03}
\end{figure*}

\noindent Fig.~\ref{fig:Appendix03} shows the qualitative results of READ on the ReasonSeg \textit{val} set. LISA and SESAME exhibit various defects to some extent when handling the displayed cases, whereas our approach delivers more desirable segmentation results.
\begin{figure*}[h]
\vspace{-0.2cm}
\begin{center}
\includegraphics[width=1\linewidth]{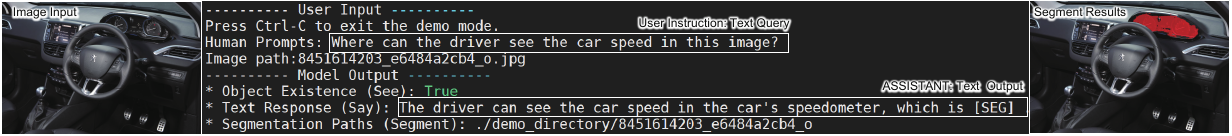}
\end{center}
\vspace{-0.6cm}
\caption{
Showcase of complex reasoning and world knowledge.}
\label{fig:mayun}
\vspace{-0.4cm}
\end{figure*}

\section{Discussion}
\label{sup:discussion}
\paragraph{Applicability.}
To showcase the broader applicability of our approach, we discuss how READ can be integrated with other methods. For LLM‐based referring segmentation, such as LISA \cite{lai2024lisa}, GSVA \cite{xia2024gsva}, and GlaMM \cite{rasheed2024glamm}, our SasP module can be seamlessly incorporated with negligible effort, as they share the same \texttt{<SEG>} token pipeline as READ (ours). For non‐LLM‐based referring segmentation, such as MMCA \cite{yao2024mmca}, we compute the similarity between the output state of the \texttt{<SEG>}-like token and the image tokens derived from the last hidden layer in transformers to obtain a similarity map. We then select highly activated points for sparse embedding representations or use these points to interpolate features from a CNN‐based (ResNet) feature map, similar to the lightweight RoI pooling operation in object detection tasks. The resulting embeddings can then be employed for downstream vision tasks. 
Beyond segmentation, as long as a vision task involves generating an attention map, our Discrete-to-Continuous (DtoC) strategy (Sec.~\ref{method:arch}) can be applied to edit the attention map.

\vspace{-0.3cm}
\paragraph{Necessity.} This raises two pivotal issues for consideration. First, is the \texttt{<SEG>} token (or a \texttt{<SEG>}-like placeholder) truly necessary? Moreover, what advantages does the \texttt{<SEG>} token offer (why \texttt{<SEG>} token)?  For the former, if the \texttt{<SEG>} token merely serves as a connector role for downstream tasks, then it is not necessary. For tasks that only involve segmenting positive samples where the object to be segmented is expected to exist (as in LISA), one could alternatively use the embeddings derived from the LLMs' output text to tap into the LLMs' capabilities. However, if the \texttt{<SEG>} token functions as a decision indicator of whether segmentation should be performed, then its inclusion becomes necessary. For instance, when it comes to false premises where the target objects might not exist, it is crucial to rely on the LLMs' prediction (specifically, whether the output contains the \texttt{<SEG>} token) to determine if segmentation should take place. 

For the latter, the \texttt{<SEG>} token infuses LLMs' world knowledge into downstream tasks, compared to non-LLM-based methods such as MMCA \cite{yao2024mmca} and M-DGT \cite{chen2022MDGT}. As illustrated in Fig.~\ref{fig:mayun}, solving the text query ``Where can the driver see the car speed?" requires the model to possess world knowledge, since the query itself does not explicitly contain semantics that point to the answer (``speedometer"). In contrast, MMCA and M-DGT use BERT and ResNet as backbones, regardless of how effective their feature embeddings are, they inherently lack additional world knowledge. 


\clearpage

\end{document}